\documentclass{article}

 \usepackage[preprint]{neurips_2026}

\usepackage[utf8]{inputenc} 
\usepackage[T1]{fontenc}    
\usepackage[hypertexnames=false]{hyperref}       
\usepackage{url}            
\usepackage{booktabs}       
\usepackage{amsfonts}       
\usepackage{nicefrac}       
\usepackage{microtype}      
\usepackage{xcolor}         

\usepackage{graphicx}
\usepackage{bm}
\usepackage{amsmath}
\usepackage{minted} 
\usepackage{algorithm}  
\usepackage[noend]{algpseudocode}
\usepackage{multirow}

\title{Computational framework for multistep metabolic pathway design\thanks{This work is Chapter 5 from P.Z.'s Ph.D. thesis \citep{zhang2021developing}.}}

\author{%
  Peter Zhiping Zhang \& Jeffrey D. Varner* \\
  Robert Frederick Smith School of Chemical and Biomolecular Engineering\\
  Cornell University\\
  Ithaca, NY 14850 \\
  \texttt{\{zz435,\: jdv27\}@cornell.edu} \\
}

\begin{document}

\maketitle

\begin{abstract}
\emph{In silico} tools are important for generating novel hypotheses and exploring alternatives in \emph{de novo} metabolic pathway design.
However, while many computational frameworks have been proposed for retrobiosynthesis, few successful examples of algorithm-guided xenobiotic biochemical retrosynthesis have been reported in the literature.
Deep learning has improved the quality of synthesis and retrosynthesis in organic chemistry applications.
Inspired by this progress, we explored combining deep learning of biochemical transformations with the traditional retrobiosynthetic workflow to improve \emph{in silico} synthetic metabolic pathway designs.
To develop our computational biosynthetic pathway design framework, we assembled metabolic reaction and enzymatic template data from public databases. 
A data augmentation procedure, adapted from literature, was carried out to enrich the assembled reaction dataset with artificial metabolic reactions generated by enzymatic reaction templates. 
Two neural network-based pathway ranking models were trained as binary classifiers to distinguish assembled reactions from artificial counterparts; each model output a scalar quantifying the plausibility of a 1-step or 2-step pathway. 
Combining these two models with enzymatic templates, we built a multistep retrobiosynthesis pipeline and validated it by reproducing some natural and non-natural pathways computationally.

\end{abstract}

\section{Introduction} \label{sec:intro}
Manufacturing molecules through metabolic engineering is key to a sustainable society, as metabolic engineering can convert renewable and widely-available starting materials into high-value products under mild manufacturing conditions \citep{keasling2010manufacturing}.
There are many successful metabolic engineering projects targeting the commercial production of bulk chemicals or pharmaceutical ingredients, such as the production of 1,4-butanediol (BDO) in \textit{E.~coli} \citep{yim2011metabolic} and the production of the antimalarial drug precursor artemisinic acid in engineered yeast \citep{ro2006production}.
However, with current technologies, the development of a new industrial bio-production process typically costs several years and millions of dollars \citep{nielsen2016engineering}. 
\textit{In silico} biosynthetic pathway design tools are important for accelerating the design-build-test-learn cycle in metabolic engineering, as they can explore the design space and generate novel synthetic pathways \citep{nielsen2016engineering, keasling2010manufacturing}.

\begin{figure}
\centering
    \includegraphics[width=\textwidth]{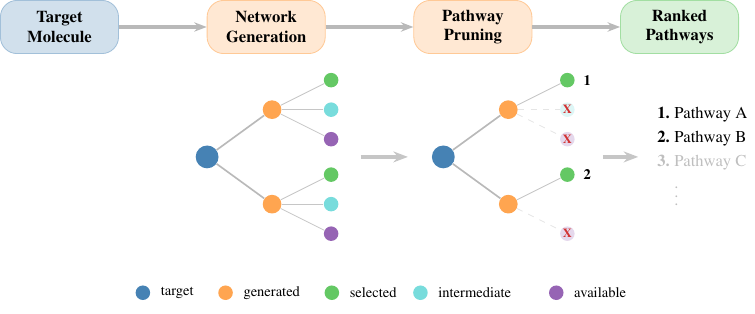}
    \caption{A general retrobiosynthesis workflow consists of two parts. In network generation, the target is used as the seed for metabolic network generation, which adopts some candidate selection algorithm to choose candidate intermediates and applies generalized enzymatic reaction rules on those intermediates to propose hypothetical reactions for network expansion.
    In pathway pruning/ranking, all generated reactions/pathways are evaluated by pruning/ranking models to reduce the number of candidates.
    All remaining pathways are extracted for qualitative and quantitative analysis. Pruning/ranking algorithms generally incorporate domain-specific knowledge, like structural similarity and enzyme availability assessment, theoretical yield in the context of a metabolic model, and thermodynamic constraints, to analyze the plausibility of each pathway. Scoring functions are popular choices for quantifying the plausibility of all remaining pathways and give a list of ranked pathway candidates for assisting experimental design. (adapted from \citep{yim2011metabolic, hadadi2015design}).}
    \label{fig:retro_workflow}
\end{figure}

Given a desired compound of interest to produce, a retrobiosynthesis tool works backward to identify a series of enzymatic transformations that can construct the target from some available cellular metabolites or a biochemical feedstock \citep{hadadi2015design, lin2019retrosynthetic}.  
Computer-assisted retrobiosynthesis tools play an important role in helping scientists explore the whole design space for generating novel \textit{de novo} biosynthetic pathways \citep{hadadi2015design}.
In analogy to chemical retrosynthesis, a general workflow for a computational retrobiosynthesis tool consists of two parts --- network generation and pathway pruning/ranking (Fig.~\ref{fig:retro_workflow}) \citep{yim2011metabolic, segler2018planning, hadadi2015design, wang2017review}. 
The target is used as the seed for metabolic network generation in a pipeline, which adopts some candidate selection algorithm to choose candidate intermediates and applies generalized enzymatic reaction rules on those intermediate candidates for network expansion. This expansion continues iteratively until reaching some stopping criteria such as a set of designated starting compounds, maximum depth of a network, or maximum allocated computational time. 
Then, all generated reactions and pathways would be evaluated by some quantitative metrics. Reactions and pathways that could not meet the set thresholds of those metrics would be removed from the generated network. 
All potential pathways from the remaining network are extracted from the generated network for qualitative and quantitative analyses. Those analyses may incorporate domain-specific knowledge, like structural similarity, enzyme availability, theoretical yield in the context of a metabolic model, and thermodynamic constraints, to analyze the plausibility of each pathway. 
In the pathway ranking step, scoring functions are designed to quantify the plausibility of all remaining pathways and produce a list of ranked pathway candidates for guiding experimental design.

Some systems have been developed for the retrobiosynthesis task. 
Developed by Genomatica, SimPheny was part of their Biopathway Predictor framework to engineer \textit{E.~coli} for the first direct biocatalytic synthesis of 1,4-Butanediol (BDO) from renewable carbohydrate feedstocks \citep{yim2011metabolic}. Metabolic models were used to analyze the performance of each proposed pathway, such as theoretical yield. 
Built upon public database Kyoto Encyclopedia of Genes and Genomes (KEGG) \citep{kanehisa1996toward, kanehisa1997database, kanehisa2000kegg, kanehisa2012kegg, kanehisa2016kegg}, the BNICE framework generated a publicly available biochemistry database ATLAS and claimed to be able to do retrobiosynthesis but is not publicly or commercially available \citep{hadadi2016atlas, hafner2020updated, hatzimanikatis2005exploring, finley2009computational, soh2010dreams, finley2010silico}.   
PathMiner provides a heuristic search method for extracting metabolic routes from a network, although it does not involve network generation \citep{mcshan2003pathminer}.
KEGG has a built-in pathway prediction server, PathPred, using SIMCOMP program and Reaction center-Different region-Matched region (RDM) pattern representation of reaction rules \citep{moriya2010pathpred, hattori2003development}. 
RetroPath proposed a data-driven approach to automatically extract reaction rules in SMARTS from biochemical reactions and incorporated enzyme sequence consistency and compound similarity into pathway ranking \citep{delepine2018retropath2}. 
They then validated this framework by using RetroPath to explore nine million possible enzyme combinations that could lead to the production of flavonoid pinocembrin and narrow it down to 12 candidates which were then used for experimental designs. 
Four out of those 12 top-ranked enzyme combinations proved to be able to produce the target compound with significant yields in the \textit{E.~coli} chassis.
RetroPathRL adopted Monte Carlo Tree Search reinforcement learning method to explore biosynthetic space for longer pathway design \citep{koch2019reinforcement}.
Despite these tools, the production of 1,4-butanediol (BDO) in \textit{E.~coli} remains the only example of a retrosynthesis algorithm being used in a \textit{de novo} pathway design for the commercial production of a xenobiotic compound \citep{lin2019retrosynthetic}. 

Due to its ability to learn knowledge representations from large amounts of data, deep learning has become increasingly popular in promoting chemical (retro)synthesis \citep{lecun2015deep, lin2019retrosynthetic}. 
Wei and coworkers combined convolutional neural network generated molecular fingerprints with a neural network for reaction type prediction to predict likely chemical reaction products given a set of reagents and reactants \citep{wei2016neural, ananiadou2010event}. 
Coley and coworkers introduced neural network based scoring function into reaction template based forward enumeration framework for predicting organic reaction outcomes and achieved $71.8\%$ accuracy in rank 1 candidate \citep{coley2017prediction}. The accuracy was improved to $85.6\%$ later by a template-free graph-convolutional neural network model \citep{coley2019graph, jin2017predicting}. 
Segler and coworkers designed a chemical retrosynthesis system (3N-MCTS) with Monte Carlo tree search and three different neural networks and achieved improved performance over traditional computer-aided search methods \citep{segler2018planning}.   


Inspired by this recent progress in computational organic chemistry, we explored the idea of combining deep learning of biochemical transformations with the traditional retrobiosynthetic workflow, to improve \emph{in silico} designs for synthetic metabolic pathways.

%

\section{Methods} \label{sec:Methods}
\textbf{Overview of the approach}. 
We adopted backward template-based enumeration for network expansion, and used deep learning based ranking models for network pruning and candidate ranking. 
To develop our computational biosynthetic pathway design framework, we assembled metabolic
reaction and enzymatic template data from public databases. A data augmentation procedure,
adapted from literature, was carried out to enrich the assembled reaction dataset with artificial
metabolic reactions generated by enzymatic reaction templates \citep{coley2017computer}. 
Two neural network-based pathway ranking models were trained as binary classifiers, by outputting a scalar quantifying the likelihood of a 1-step or 2-step pathway being plausible, to distinguish assembled reactions from artificial counterparts. Combining these two models with enzymatic templates, we built a
multistep retrobiosynthesis pipeline and validated it by reproducing some natural and non-natural pathways computationally.

\subsection{Dataset assembly}
The backward reaction templates and metabolic reactions were assembled from literature \citep{henry2010discovery, duigou2019retrorules} and public databases \citep{kanehisa2016kegg, EPFL_ATLAS, hadadi2016atlas}. 

\textbf{Metabolic reaction dataset.} KEGG (Kyoto Encyclopedia of Genes and Genomes) is a computerized resource for understanding high-level functions and utilities of the biological system \citep{kanehisa2016kegg, kanehisa1996toward, kanehisa1997database, ogata1999kegg, kanehisa2000kegg, kanehisa2002kegg, kanehisa2004kegg, kanehisa2006genomics, kanehisa2007kegg, kanehisa2010kegg, kanehisa2012kegg, kanehisa2019toward, kanehisa2019new, kanehisa2017kegg, kanehisa2014data, kanehisa2016kegg}
It is a collection of databases dealing with many aspects of information in biology, including biological pathways and chemical substances \citep{kanehisa2016kegg}. 
We assembled 11475 reactions from KEGG Reaction Database and 323 pathways from KEGG Module Database. 
Each reaction has its unique KEGG reaction ID, corresponding enzyme as EC number, and compound names as KEGG compound ID. 
The dataset was modest in size, but our purpose here was to build a working pipeline as a proof-of-concept study.
Additional data would improve the performance of the pipeline developed here. 
A better way to resolve dataset issue is to assemble more data directly from various databases, such as ChEBI \citep{chebi}, XTMS \citep{xtms}, Rhea \citep{rhea, morgat2016updates}, and MINEs. 

\textbf{Backward reaction templates dataset.}
Building a pipeline for reaction template extraction was not a trivial job, so instead in this work, we decided to assemble a backward reaction templates dataset from literature. 
Faulon group published multiple sets of reaction rules based on metabolic reaction database MetaNetX v3.0 \citep{duigou2019retrorules, metanetx, moretti2016metanetx, ganter2013metanetx, bernard2014reconciliation}. 
We used their reaction rules dataset that handles hydrogen explicit, and with a template diameter from 2 to 16 \citep{duigou2019retrorules}.
The dataset had 350224 templates, including both forward and backward ones. 
Among them, there were 234268 backward templates, which would be referred to as 'RetroRule' rule set.  
We also obtained a dataset that was used in Henry et al.\ \citep{henry2010discovery} that consisted of 116 backward rules, referred to as the `BNICE' rule set in this paper.

\subsection{Baseline model}
We developed a baseline model on top of the Tanimoto similarity metric as elaborated in Section~\ref{sec:SI-PR}.
Two backward rule sets, RetroPath and BNICE, were applied on product(s) of each positive reaction to generate negative derivatives, and then Tanimoto scores were calculated with each reaction represented as a 1024-bit vector as discussed in Section~\ref{sec:SI-methods-RF}.
The final ranks of positive reactions were solely based on Tanimoto scores. 

\subsection{Neural Network based 1-step Pathway Ranking model (NN1PR)}
\begin{figure}
\centering
    \includegraphics[width=\textwidth]{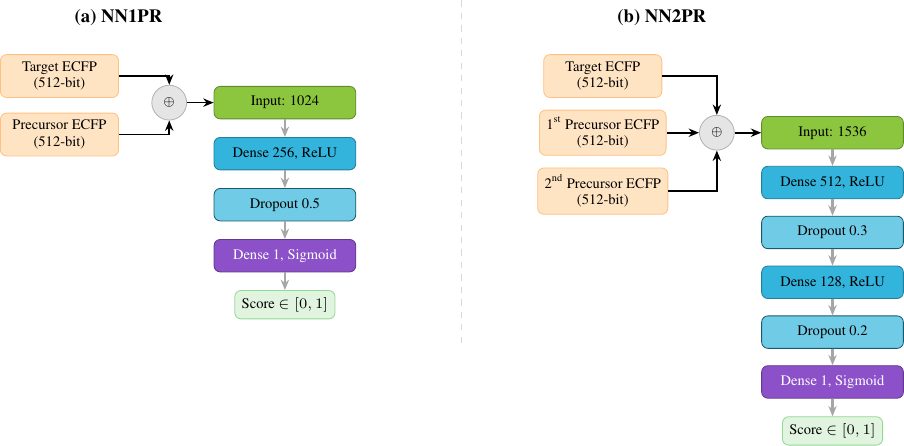}
    \caption{Architectures of the neural network pathway ranking models. (a)~NN1PR: a 1-step pathway ranking model that takes the concatenation of target and precursor ECFPs (1024-bit) as input. (b)~NN2PR: a 2-step pathway ranking model that takes the concatenation of target, first-step precursor, and second-step precursor ECFPs (1536-bit) as input. Both models output a scalar score $\in [0,1]$ quantifying the plausibility of the pathway.}
    \label{fig:model_architectures}
\end{figure}

A multilayer perceptron (MLP) model, consisting of 1 hidden layer with 256 neurons and a dropout layer, was trained as a binary classifier to distinguish assembled reactions from their
generated counterparts (Fig.~\ref{fig:model_architectures}a). 
The input to the network was the concatenation of ECFPs of target molecule and precursor candidate.
The final scalar output from the trained model was then used as a quantitative likelihood measure of a reaction being plausible.
For comparison, we developed a baseline model based on Tanimoto similarity \citep{tanimoto1958elementary}.
The score for each target-precursor pair was given by the Tanimoto similarity between the target and proposed precursor.
Details on this baseline model are discussed in Section~\ref{sec:SI-PR}.
Both the baseline model and NN1PR model were used to rank positive reactions, which were one-step metabolic reactions from KEGG, among their corresponding negative counterparts generated through a data augmentation procedure as discussed in Section~\ref{sec:SI-method-DA}. 
About 10\% assembled KEGG reaction data were reserved for performance comparison. Assembled data were randomly shuffled before training/testing split to guarantee a fair coverage in both training and testing datasets.

\subsection{Neural Network based 2-step Pathway Ranking model (NN2PR)}

To incorporate more information from previous steps when ranking a pathway, another neural network model was developed.
A multilayer perceptron (MLP) model, consisting of two hidden layers with 512 and 128 neurons separately and two dropout layers, was trained as a binary classifier to distinguish assembled 2-step pathways from their generated counterparts (Fig.~\ref{fig:model_architectures}b). 
The input to the network was the concatenation of ECFPs of target molecule, first-step precursor candidate and second-step precursor. 
The final scalar output from the trained model was then used as a quantitative likelihood measure of a 2-step pathway being plausible.
NN1PR was used as the baseline of NN2PR for performance comparison. 
When ranking a 2-step pathway, NN1PR used only information from the most recent step, while NN2PR used information from both steps. 
Both NN2PR and NN1PR models were used to rank 2-step pathways, among their corresponding negative counterparts generated through a data augmentation procedure as discussed in Section~\ref{sec:SI-method-DA}. 
About 10\% assembled KEGG pathway data were reserved for performance comparison. Assembled data were randomly shuffled before training/testing split to guarantee a fair coverage in training and testing data set.


\subsection{Pathway design framework with NN1PR}
Combining template-based backward enumeration with NN1PR, we built a pipeline for one-step retrosynthetic reaction design \citep{coley2017computer}.
In the backward enumeration, the selected template set was applied on the given target for retrobiosynthesis to generate precursor candidates $R_1, R_i, \ldots, R_n$.
Together with the target $T$, they formed target-precursor pairs ($T$--$R_1$, $T$--$R_i$, \ldots, $T$--$R_n$), that were then fed to NN1PR which would assign a scalar score in unit range to each pair.
Using NN1PR as the ranking model, together with enzymatic template-based network generation, we also built a multistep retrobiosynthesis pipeline prototype.
Given a compound for retrobiosynthesis, enzymatic templates were applied for 1st-step backward enumeration, followed by NN1PR ranking. High-ranked candidates were chosen for 2nd-step enumeration and results were ranked again by NN1PR. The process continued until reaching the specified maximum step. A pathway reconstruction step would then search through all generated pathways for desired ones.

\subsection{Multistep pathway design framework with NN1PR and NN2PR}
\begin{figure}
\centering
    \includegraphics[width=\textwidth]{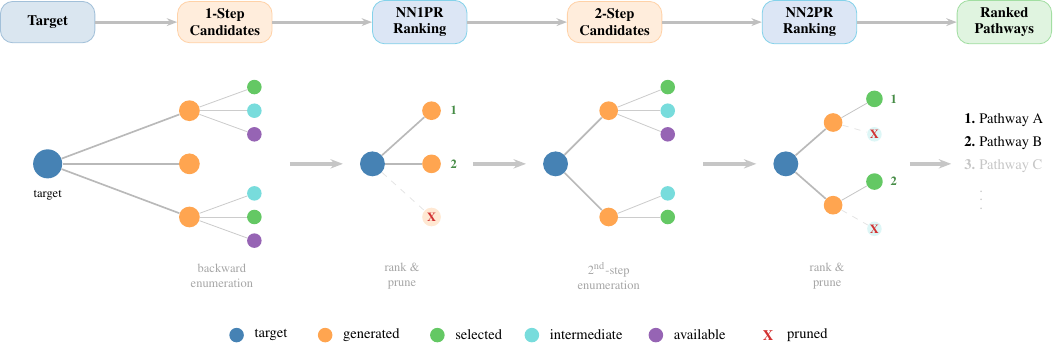}
    \caption{Proposed multistep metabolic pathway design pipeline with NN1PR and NN2PR as ranking models.
    Given a compound for retrobiosynthesis, enzymatic templates are applied for 1st-step backward enumeration, followed by NN1PR ranking. High-ranked candidates are chosen for 2nd-step enumeration and results are pruned by NN1PR before being ranked by NN2PR. The process goes on as in the 2nd step until reaching the specified maximum step. A pathway
    reconstruction step would then search through all generated pathways for desired ones (not shown in the figure).}
    \label{fig:multistep_pipeline}
\end{figure}
Combining NN1PR and NN2PR with enzymatic templates, we improved our
multistep retrobiosynthesis pipeline and validated it through reproducing some natural and non-natural pathways computationally (Fig.~\ref{fig:multistep_pipeline}).
Given a compound for retrobiosynthesis, enzymatic templates were applied for 1st-step backward enumeration, followed by NN1PR ranking. High-ranked candidates were chosen for 2nd-step enumeration and results were pruned by NN1PR before being ranked by NN2PR. The process went on as in the 2nd step until reaching the specified maximum step. A pathway reconstruction step would then search through all generated pathways for desired ones.

%
\section{Results} \label{sec:results}
\begin{figure}
\centering
    \includegraphics[width=\textwidth]{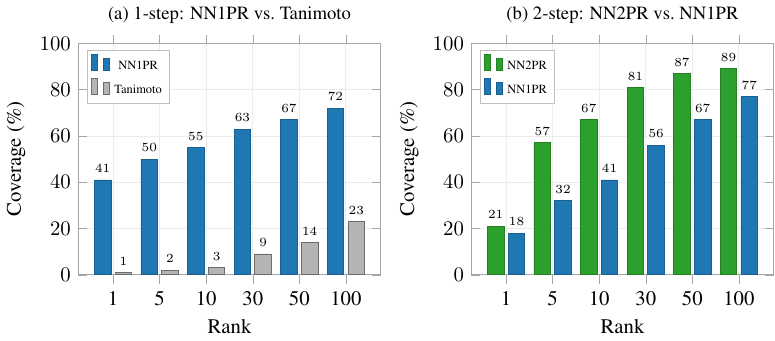}
    \caption{Performance comparison on testing data. (a)~1-step ranking: NN1PR vs.\ Tanimoto similarity baseline. About 55\% of samples were ranked among the top-10 by NN1PR, compared to only 3\% by the baseline. (b)~2-step ranking: NN2PR vs.\ NN1PR. NN2PR increased top-10 coverage by about 26\% over NN1PR, but the improvement was narrower as rank increased to top-100.}
    \label{fig:model_performance}
\end{figure}

NN1PR outperformed its baseline model (Fig.~\ref{fig:model_performance}a).
About 55\% of samples were ranked among the top-10 by NN1PR, compared to only 3\% by the Tanimoto baseline. Top-100 candidates by NN1PR covered about 72\% of samples, showing room for improvement.
In contrast, to reach 90\% coverage, the baseline had to go higher than $1000$ in rank.
NN2PR increased top-10 coverage by about 25\% over NN1PR, but the improvement narrowed as rank increased (Fig.~\ref{fig:model_performance}b).
Increasing top-10 coverage was useful for longer pathway design as it reduced combinatorial explosion as the backward expansion extended.
The better performance of NN2PR over NN1PR showed the benefit of incorporating more pathway information in pathway ranking.

Having established the ranking performance of both models on held-out data, we next assessed whether the one-step framework could recover known biosynthesis pathways.
We validated the framework by reproducing the BDO pathway reported by Yim et al.\ \citep{yim2011metabolic}.
Given the target, 1,4-Butanediol (BDO), for biosynthesis, we applied the framework with only the BNICE rule set (116 templates) \citep{henry2010discovery, duigou2019retrorules} and found the desired precursor 4-Hydroxybutyraldehyde with EC $1.1.1.-$ at rank 9 out of 9 candidates.
For the next step, the pipeline expanded on 4-Hydroxybutyraldehyde to generate precursors and found 4-Hydroxybutyryl CoA with EC $3.1.2.-$, ranked 1 out of 23 alternatives.
The network expansion continued for another 4 steps to completely reproduce the reported BDO pathway.
Note that the BDO pathway is a non-natural pathway designed by human experts.
Reproducing the BDO pathway showed that this framework, although trained on natural metabolic data, could generate non-natural pathways with high ranks.

\begin{figure}
\centering
    \includegraphics[width=0.9\textwidth, trim={0cm, 0cm, 0cm, 0cm}, clip]{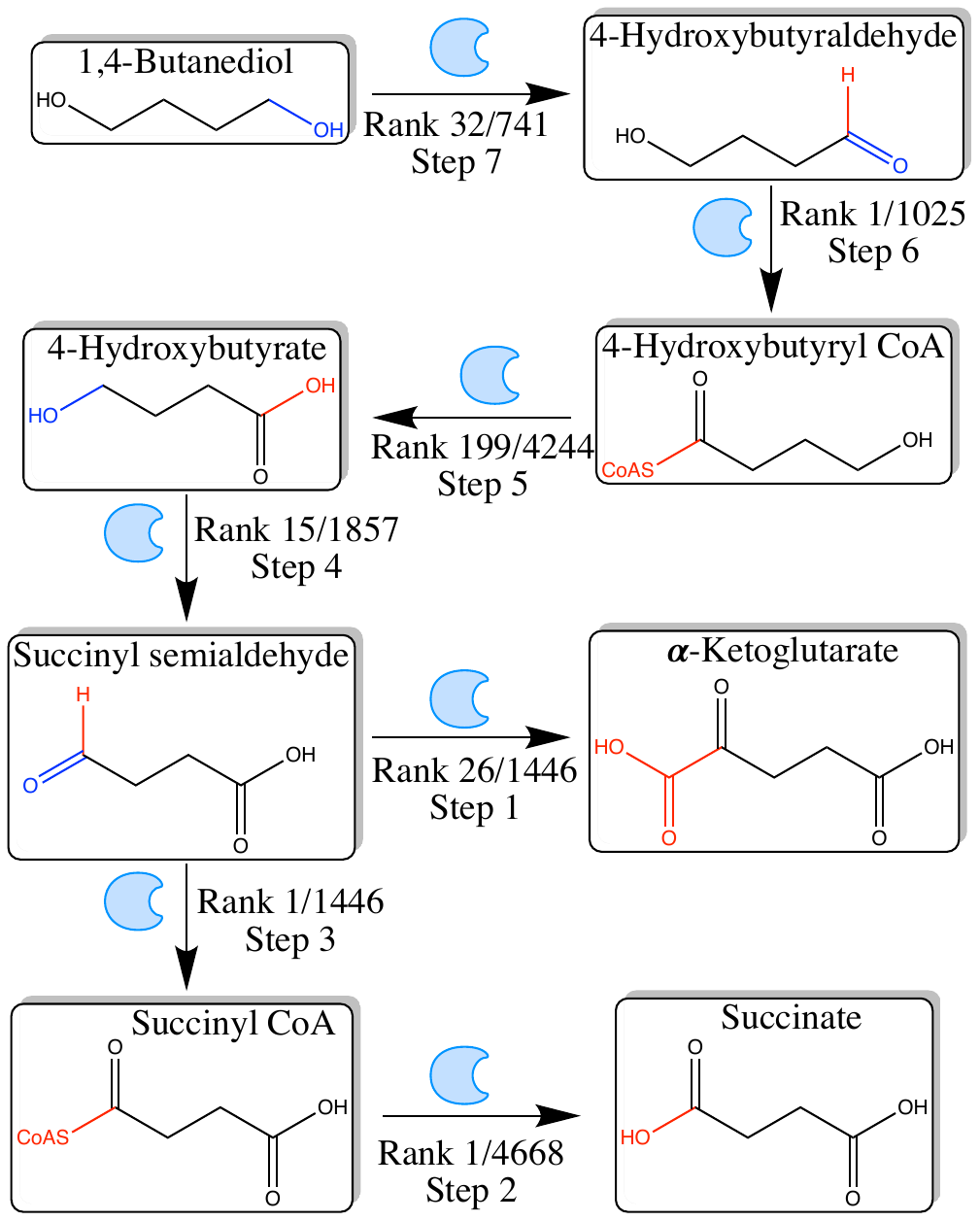}
    \caption{Ranks assigned by NN1PR to the BDO pathway reported in \citep{yim2011metabolic} in a backward manner with 234501 templates (116 BNICE + 234384 RetroPath). Each step is annotated with a rank. Multiple ECs could produce the desired result in each step as summarized in Table~\ref{table:ECs_BDO}.}
    \label{fig:bdo_pathway}
\end{figure}

To see how our framework would perform as the size of the template set scaled up, we added another 234384 templates from RetroPath and reran our pipeline (Fig.~\ref{fig:bdo_pathway}).
The number of templates jumped up by about 2021 times, from 116 to 234500, and the number of candidates increased by 71 times on average, but the absolute rank was at most 4 times worse (rank of 1,4-butanediol to 4-hydroxybutyraldehyde increased from 9 to 32).
This comparison showed the framework scaled well with larger template sets.

\begin{table}[ht]
\centering
\caption{EC numbers for BDO pathway using 234501 reaction templates.}
\label{table:ECs_BDO}
    \begin{tabular*}{\linewidth}{c|p{13cm}}
    \toprule
    {\bf Step} & {\bf EC numbers} \\ 
    \midrule
    1 & 4.1.1.79, 4.1.1.71, 4.1.1.1, 2.6.1.19, 4.1.1.74, 4.1.1.72, 4.1.1.40, 4.1.1.43, 4.1.1.80, 4.1.1.7, 4.1.1, 4.1.1.75, 4.1.1.-, 4.1.1.82 \\
    2 & 2.8.3.8, 2.8.3.1, 2.8.3.14, 2.8.3.18, 2.8.3.12, 2.8.3.- \\
    3 & 3.1.2.- \\
    4 & 1.1.1.19, 1.1.1.72, 1.1.1.317, 1.1.1.10, 1.1.1.126, 1.1.1.224, 1.1.1.2, 1.1.1.131, 1.1.1.153, 1.1.1.365, 1.1.1.307, 1.1.1.50, 1.1.1.188, 1.1.1.60, 1.1.1.194, 1.1.1.265, 1.1.1.324, 1.3.1.20, 1.1.1.137, 1.1.1.216, 1.1.1.213, 1.1.1.3, 1.1.1.1, 1.1.1.156, 1.1.1.372, 1.1.1.51, 1.1.1.62, 1.1.1.79, 1.1.1.105, 1.1.1.273, 1.1.1.91, 1.1.1.33, 1.1.1.20, 1.1.1.21, 1.1.1.184, 1.1.1.149, 1.1.1.209, 1.1.1.151, 1.1.1.55, 1.1.1.298, 1.1.1.71, 1.1.1.313, 1.1.1.54, 1.1.1.177, 1.1.1.283, 1.1.1.-, 1.1.1.195, 1.1.1.65, 1.1.1.200, 1.1.1.300, 1.1.1.191 \\
    5 & 2.8.3.8, 2.8.3.1, 2.8.3.14, 2.8.3.18, 2.8.3.12, 2.8.3.- \\
    6 & 3.1.2.- \\
    7 & 1.1.1.19, 1.1.1.72, 1.1.1.317, 1.1.1.10, 1.1.1.126, 1.1.1.224, 1.1.1.2, 1.1.1.131, 1.1.1.153, 1.1.1.365, 1.1.1.307, 1.1.1.50, 1.1.1.188, 1.1.1.60, 1.1.1.194, 1.1.1.265, 1.1.1.324, 1.3.1.20, 1.1.1.137, 1.1.1.216, 1.1.1.213, 1.1.1.3, 1.1.1.1, 1.1.1, 1.1.1.156, 1.1.1.372, 1.1.1.51, 1.1.1.62, 1.1.1.79, 1.1.1.105, 1.1.1.273, 1.1.1.91, 1.1.1.33, 1.1.1.20, 1.1.1.21, 1.1.1.184, 1.1.1.149, 1.1.1.209, 1.1.1.151, 1.1.1.55, 1.1.1.298, 1.1.1.71, 1.1.1.313, 1.1.1.54, 1.1.1.177, 1.1.1.283, 1.1.1.-, 1.1.1.195, 1.1.1.65, 1.1.1.200, 1.1.1.300, 1.1.1.191 \\
    \bottomrule
	\end{tabular*}
\end{table}

\begin{figure}
\centering
    \includegraphics[width=0.9\textwidth, trim={0cm, 0cm, 0cm, 0cm}, clip]{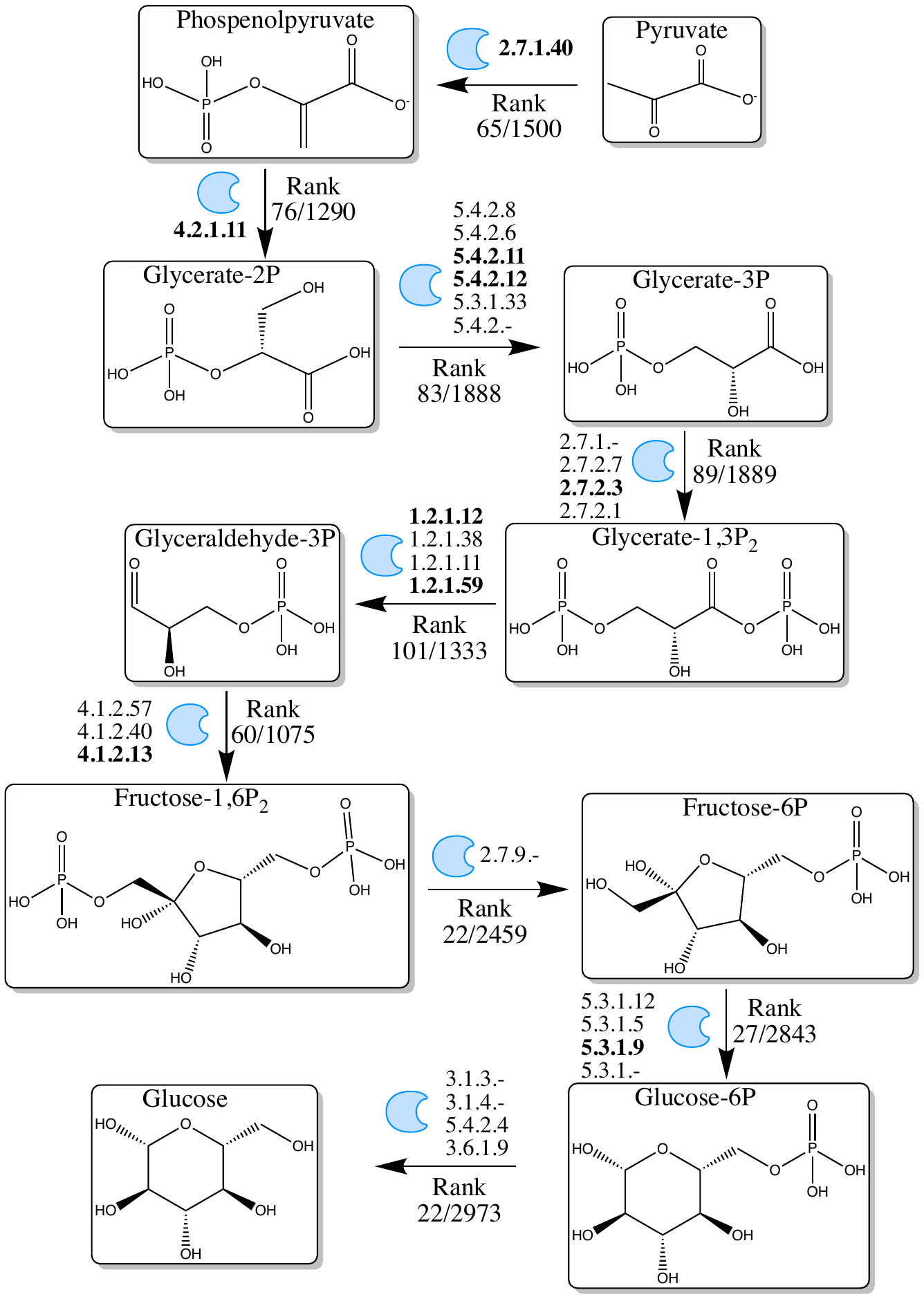}
    \caption{Ranks assigned by NN1PR for glycolysis pathway in a backward manner. 234501 templates were used in backward enumeration step to produce results here. Ranks here were the best ranks if there were alternatives leading to the same target. ECs were ECs associated with the corresponding rank, and ECs in bold were the ECs recorded in KEGG.}
    \label{fig:pyruvate_pathway}
\end{figure}

We also applied the one-step framework to a natural pathway by reproducing glycolysis.
Given pyruvate as the target for biosynthesis, we applied the framework with the BNICE and Retro rule sets \citep{duigou2019retrorules} and found the desired precursor phosphoenolpyruvate with EC $2.7.1.40$, as reported in KEGG, at rank 65 out of 1500 candidates (Fig.~\ref{fig:pyruvate_pathway}).
The network expansion went on repeatedly for another 8 times to completely reproduce the well-known glycolysis pathway. 
For 7 steps in this reproduction, the recorded ECs in KEGG were included in the best ranked candidates found by the pipeline. 
While for the other two steps (fructose 1,6-bisphosphate to fructose 6-phosphate and glucose 6-phosphate to glucose), the recorded ECs were included in other alternative candidates.  

\begin{figure}
\centering
    \includegraphics[width=\textwidth, trim={0cm, 0cm, 0cm, 0cm}, clip]{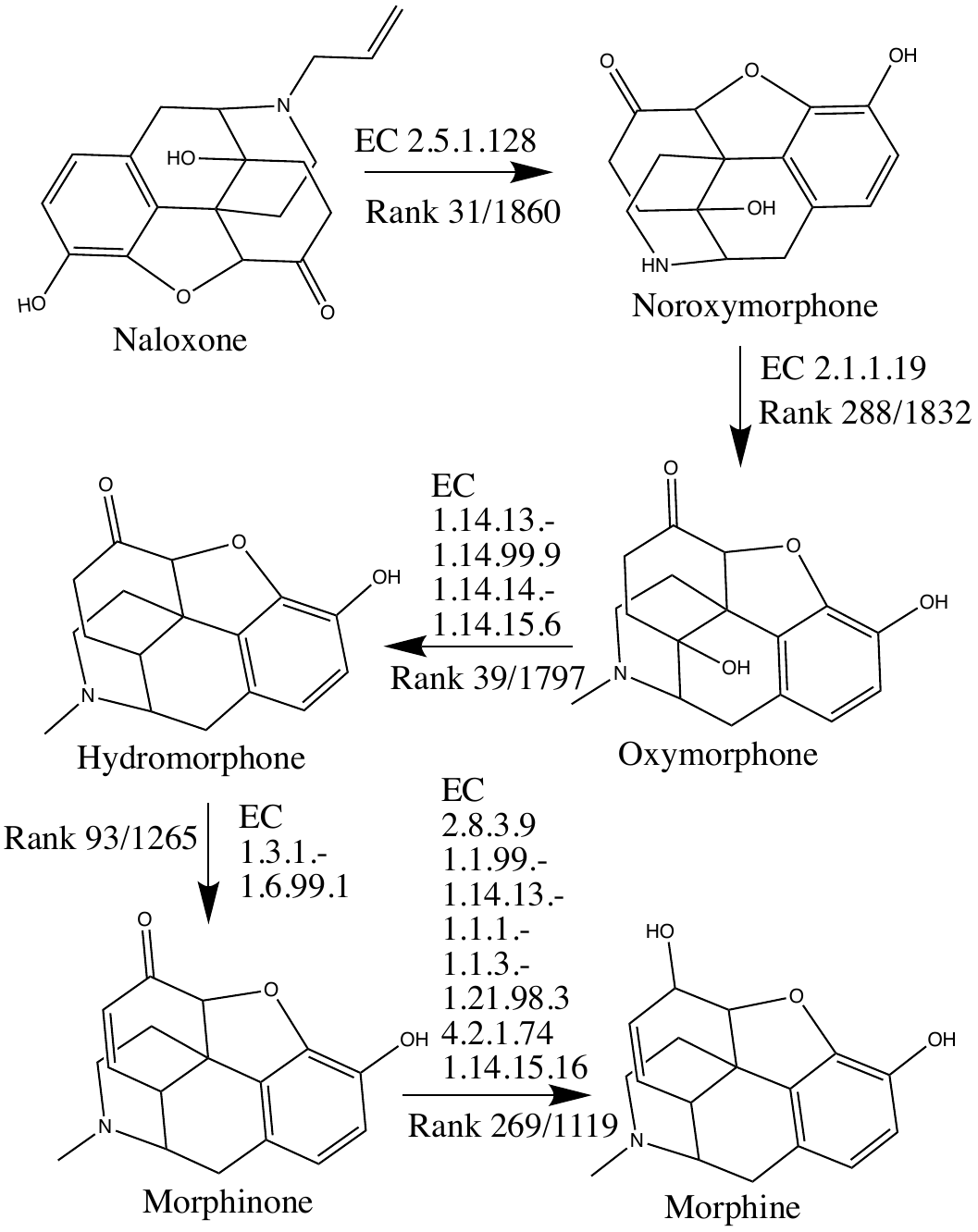}
    \caption{Ranks assigned by NN1PR for the naloxone pathway designed by human expert in a backward manner. Each step was annotated with a rank and corresponding EC number(s). The 116 templates in BNICE rule set and 234384 templates from RetroPath were used in backward enumeration step to produce results in this figure.}
    \label{fig:morphine_pathway}
\end{figure}
We also evaluated a human-designed naloxone synthesis pathway proposed by a colleague (Fig.~\ref{fig:morphine_pathway}).
Some steps had multiple matched candidates, while others had only one; only the best ranks and corresponding ECs were shown.
The ranks evaluated by the pipeline provided a quantitative understanding of the difficulty in achieving each step, and the corresponding ECs guided the identification of candidate enzymes.

\begin{figure}
\centering
    \includegraphics[width=0.98\textwidth, trim={3cm, 5cm, 2cm, 5cm}, clip]{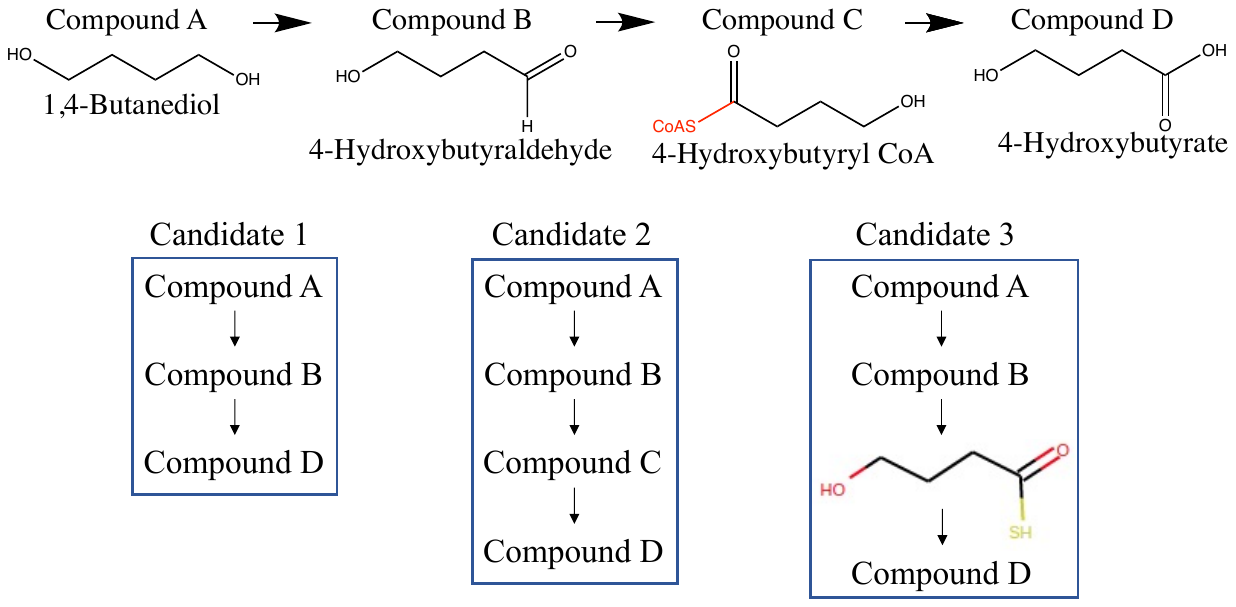}
    \caption{Three candidates for BDO synthesis found by the proposed pipeline. At the top is the reported pathway that synthesizes BDO (1,4-Butanediol) from 4-Hydroxybutyrate. Three candidates were found by the computational pipeline proposed. Candidate 1 skipped compound C in the original pathway; candidate 2 was an exact match; candidate 3 replaced the -CoA group in compound C with -SH group.}
    \label{fig:BDO_candidates}
\end{figure}

Beyond step-by-step analysis, we tested the fully automated multistep pipeline, first using NN1PR alone as the ranking model.
We validated this pipeline by reproducing the BDO pathway computationally \citep{yim2011metabolic}.
When running the pipeline on the production of 1,4-BDO from 4-Hydroxybutyrate, it proposed $49806$ pathways but only three candidate pathways fulfilled the assigned task (Fig.~\ref{fig:BDO_candidates}).
Compared to the original pathway, candidate 1 skipped one step by bypassing 4-Hydroxybutyryl CoA, but received a rank of $165$ out of about $50k$ candidates.
Candidate 2 was an exact match as reported in the literature but was ranked as $33662$, or around 67\% among all candidates, which was lower than expected.
Candidate 3 replaced the \textit{-CoA} functional group with a thiol group \textit{-SH}, making this pathway difficult, if not impossible, to achieve in metabolism, even though its rank, $16572$, was better than candidate 2.

We then evaluated the improved multistep pipeline combining NN1PR and NN2PR.
For the non-natural pathway validation, we ran this pipeline on 4-Hydroxybutyrate for the production of 1,4-BDO, repeating the same experiment as the NN1PR-only pipeline.
As the network generation part of this pipeline was the same as the previous one, the candidates found were the same (Fig.~\ref{fig:BDO_candidates}), but their ranks were improved.
Candidate 1 was ranked as $122$ out of about $50k$ candidates and candidate 3 got a rank of $10484$. 
For candidate 2, its rank improved significantly to $262$, which was higher than candidate 3 and hence more consistent with our knowledge in metabolism. 
In terms of rank in percentage, candidate 2 was ranked in the top $0.5\%$, compared to $67\%$ in the previous pipeline. 
This improvement showed the benefit of NN2PR for computational multistep metabolic pathway design.

For the natural pathway validation, the pipeline was given fructose 1,6-bisphosphate as the starting compound and explored backward to determine if it could recover the path to glucose.
For the step from fructose 6-phosphate to fructose 1,6-bisphosphate, the pipeline was able to recover all three ECs (2.7.1.90, 2.7.1.146, and 2.7.1.11) as in KEGG. 
The EC (5.3.1.9) for the conversion between fructose 6-phosphate and glucose 6-phosphate was also included in the prediction result. 
From glucose to glucose 6-phosphate, there were five ECs (2.7.1.1, 2.7.1.147, 2.7.1.63, 2.7.1.2, and 2.7.1.199) recorded in KEGG, but the pipeline only recovered the first two, even though missing the last one was expected as it was used for extracellular glucose. 

%
\section{Discussion and Future Direction} \label{sec:discussion}

The proposed pipelines provided a way of leveraging big-data in metabolism to facilitate metabolic pathway design by combining template-based backward enumeration with deep learning based ranking models.
However, there were several limitations, and improvements could be made to the efficiency, capability, and bio-feasibility of the pipeline.

The performance of the pipeline depended heavily on the quality of its template set, but there was no standard way of quantifying the quality of such a dataset, and it was also difficult to pre-select a template set for designing an unknown pathway.
Inserting a module that could extract templates from metabolic reactions, rather than assembling them from literature, could further automate the pipeline.
There have been some works in chemical synthesis that practiced this idea \citep{coley2017prediction, coley2017computer, segler2018planning}, but template extraction is much harder in biochemistry given that many reactions in public databases are usually not atom-balanced.
Faulon et al.\ published a work on extracting enzymatic templates, but their methodology was too constrained, resulting in conservative templates \citep{duigou2019retrorules}.
Runtime was another concern.
A good and flexible template extraction module could improve the capability of the pipeline, but runtime would become a concern as more templates became available.
One potentially effective way of reducing running time would be to develop a template selection model to screen all templates and pick out only the most suitable ones to apply on a target.
Such models should guarantee faster screening for each template than running it on the target; otherwise, the runtime benefit may not be realized.
Segler and Waller developed such a model using a neural network, although details and runtime of this model were not reported \citep{segler2018planning}.
Neural network models are fast at prediction, which makes them a good choice for template selection \citep{goodfellow2016deep}.

Beyond template selection, the demonstrated application on multiple-step design only utilized one-step and two-step ranking models, which did not take more available information into consideration.
To improve performance, neural network based n-step pathway ranking (NNnPR) models could be developed to rank n-step pathways.
Knowledge from metabolism should be incorporated to rule out common precursor candidates at early stages, so that the pipeline could focus more on exploring the unknown.
Many candidates differed from each other only by a trivial co-substrate that did not impact further backward exploration biologically; the pipeline could be improved by merging those candidates together to focus backward enumeration on main substrates.
A practical way to achieve this would be to create knowledge-based lists of equivalent substrates and use these lists to rule out equivalent candidates in network generation.
Monte-Carlo tree search could also be explored for generating longer pathways if available metabolites are given \citep{koch2019reinforcement}.

A promising way of addressing both capability and runtime issues would be to adopt template-free methods \citep{coley2019graph, jin2017predicting}.
Instead of using templates, this type of framework tries to predict what bonds and atoms in a target will change in a reaction, and then enumerate all possible outcomes.
Jin and Coley developed a graph-convolutional neural network model that had comparable performance with template-based models \citep{jin2017predicting}.
Transfer learning has been shown to be an effective approach when large datasets are not available \citep{pan2009survey}.
Considering the many advances in chemical synthesis, transfer learning could be explored for biochemical synthesis.
One possibility would be to adapt learned fingerprint models using convolutional networks to generate more informative fingerprints for biochemical compounds \citep{duvenaud2015convolutional, gomez2018automatic, wei2016neural}.

Finally, in this work only substructure information encoded as fingerprints was incorporated for ranking, but there were more quantitative structure-attribute relations that should be considered.
Metabolic reactions happen inside a chassis that contains a complicated biological environment, including pH and other substrates.
All of these factors impact the progress of each reaction to varying degrees, but were not incorporated into the pipeline.
To make this kind of model more helpful for metabolic pathway design, additional information on enzyme design and genome sequences should be incorporated \citep{keasling2010manufacturing}.
Models for evaluating the synthesizability and toxicity of a molecule could be incorporated in network generation to prune out undesirable molecules \citep{gao2020synthesizability, jang2020structure}.
Thermodynamic constraints, like Gibbs free energy, should be considered for evaluating a reaction candidate \citep{yim2011metabolic}, and pathway yield and compatibility in a chassis could be analyzed for pathway selection if a chassis is specified \citep{delepine2018retropath2}.

%
\section{Conclusion}
Combining backward template-based enumeration with neural network based ranking models, we developed a new framework for computer-aided metabolic pathway retrosynthesis. 
The deep learning ranking models were trained on KEGG metabolic reactions and outperformed the Tanimoto similarity-based method. 
The framework was demonstrated to be suitable for multistep pathway design by reproducing one natural and one non-natural pathway. 
The pipeline had several limitations, especially dependence on the quality of templates and runtime concerns, which were discussed above.
While experimental validation of the predictions remained a work in progress, many directions were identified for improving the accuracy, capability, and efficiency of the multistep computational pipeline.
This work showed the utility of deep learning for metabolic pathway design.

\section{Acknowledgements}
The work was supported by the Center on the Physics of Cancer Metabolism at Cornell University through Award Number 1U54CA210184-01 from the National Cancer Institute. The content is solely the responsibility of the authors and does not necessarily represent the official views of the National Cancer Institute or the National Institutes of Health.
We also acknowledge the financial support to J.V. and P.Z. from the departmental funding of the Robert Frederick Smith School of Chemical and Biomolecular Engineering, Cornell University.


\bibliography{Reference}

\appendix

\setcounter{figure}{0}
\renewcommand{\thefigure}{S\arabic{figure}}
\renewcommand{\theHfigure}{S\arabic{figure}}
\setcounter{table}{0}
\renewcommand{\thetable}{S\arabic{table}}
\renewcommand{\theHtable}{S\arabic{table}}
\setcounter{section}{0}
\renewcommand{\thesection}{S\arabic{section}}
\renewcommand{\thesubsection}{S\arabic{section}.\arabic{subsection}}
\renewcommand{\thesubsubsection}{S\arabic{section}.\arabic{subsection}.\arabic{subsubsection}}

\section{Supplementary Information} \label{sec:SI}



\subsection{Manipulating molecules in cheminformatics}
\textbf{SMILES and molecules.}
Representing molecules in a computer-understandable format is the first step towards manipulating molecules computationally. 
Line notation has been studied for several decades as a way of chemical nomenclature that can be processed with relative ease by computer but is also relatively understandable by human beings compared to other formats like connection table blocks \citep{panico1993guide, wiswesser1954line, dalby1992description}. 
Based on principles of molecular graph theory, David Weininger and colleagues in 1980s proposed SMILES (Simplified Molecular Input Line Entry System) as a chemical notation for modern chemical information processing \citep{weininger1988smiles, weininger1989smiles}.   
With further developments over the past several decades, mainly by Daylight Chemical Information Systems, SMILES is now widely supported by many applications as standard input and/or output formats \citep{weininger1990smiles, steinbeck2003chemistry, o2011open, daylightsmiles}. An open standard called OpenSMILES was published in the open-source chemistry community in 2007 to further support the use of SMILES \citep{opensmiles, weininger1988smiles}.

\begin{figure}
\centering
    \includegraphics[width=\textwidth, trim={6cm, 9.5cm, 6cm, 6cm}, clip]{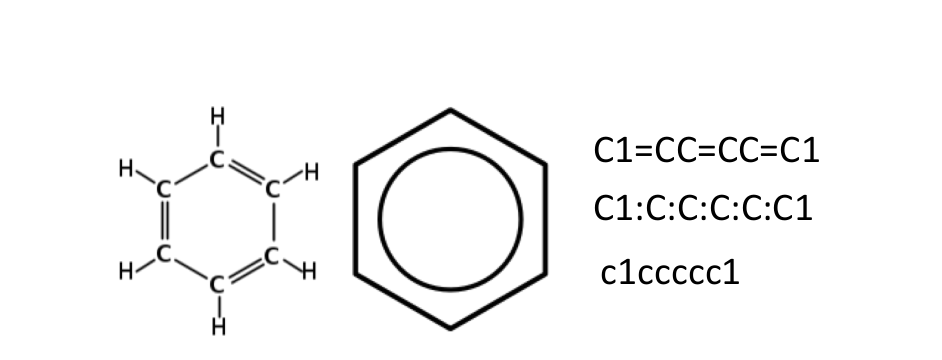}
    \caption{Demonstrating aromaticity specification with benzene.
    On the left is benzene's Kekul\'e form with alternating single and double bonds, while in the middle is another way of representing its 2D structure
    \citep{kekuie1866untersuchungen, kekule1865constitution}. 
    Three SMILES strings on the right demonstrate three ways of specifying aromaticity in SMILES corresponding to Kekul\'e, aromatic bond symbol, and atoms in lower-case forms, respectively. Kekul\'e form corresponds to Kekul\'e 2D structure, while the other two correspond to the same 2D structure in the middle.}
    \label{fig:stereo_ex}
\end{figure}

A SMILES string is read from left to right to generate atoms and bonds of a molecule. Some basic specification rules in SMILES that are important to retrosynthesis in metabolic pathway design include \citep{weininger1988smiles, daylightsmiles}: 
\begin{itemize}
    \item Atoms: representing chemical elements, such as C, N, O, P, S. 
    \item Bonds: single bond as $'-'$, double bond as $'='$, triple bond as $'\#'$, aromatic bond as $':'$.
    \item Rings: one bond in each ring is broken to maintain line notation.
    \item Aromaticity: in Kekul\'e form, or using the aromatic bond symbol, or denoting atoms in lower-case, as illustrated in Fig.~\ref{fig:stereo_ex}. 
\end{itemize}
SMILES is not without limitations. 
SMILES by its own definition does not have a canonical format, but software implementations usually provide an algorithm for standardization \citep{rdkit, willighagen2017chemistry}.
Some limitations of SMILES stem from the approximation of molecules as static graphs with well-defined bond orders. These include the inability to recognize that two molecules are tautomers of each other \citep{lowe2012extraction, carbonell2013stereo}.
In this work, SMILES was chosen as the molecular representation because of the support by many excellent software packages, especially RDKit which will be introduced later \citep{rdkit}.

\textbf{SMARTS and reaction templates.} SMILES arbitrary target specification (SMARTS) is a language for specifying substructural patterns using rules that are straightforward extensions of SMILES. Almost all SMILES specifications are valid SMARTS targets \citep{daylightsmarts}.  
Two distinctive features of SMARTS are the supports of specifying atomic properties - such as atom-atom mapping (AAM) - and logical operators for combing atom and bond descriptors. 
\begin{figure}
\centering
    \includegraphics[width=\textwidth, trim={1cm, 8cm, 4cm, 4.7cm}, clip]{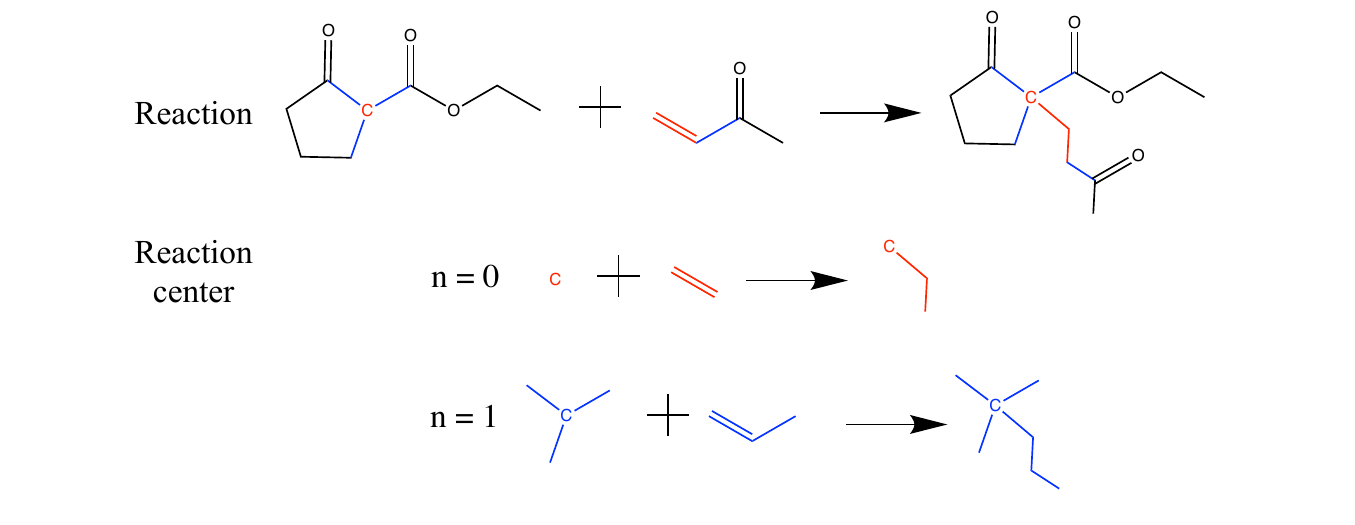}
    \caption{An example demonstrating the definition of generic reaction templates based on the distance (n) to the reaction center (adapted from \citep{lin2019retrosynthetic}). 1st row is the original reaction, and the other are two different reaction templates derived from this reaction with increasing specificity.}
    \label{fig:reaction_template_ex}
\end{figure}
\begin{figure}
\centering
    \includegraphics[width=0.85\textwidth, trim={1cm, 0cm, 1cm, 0.5cm}, clip]{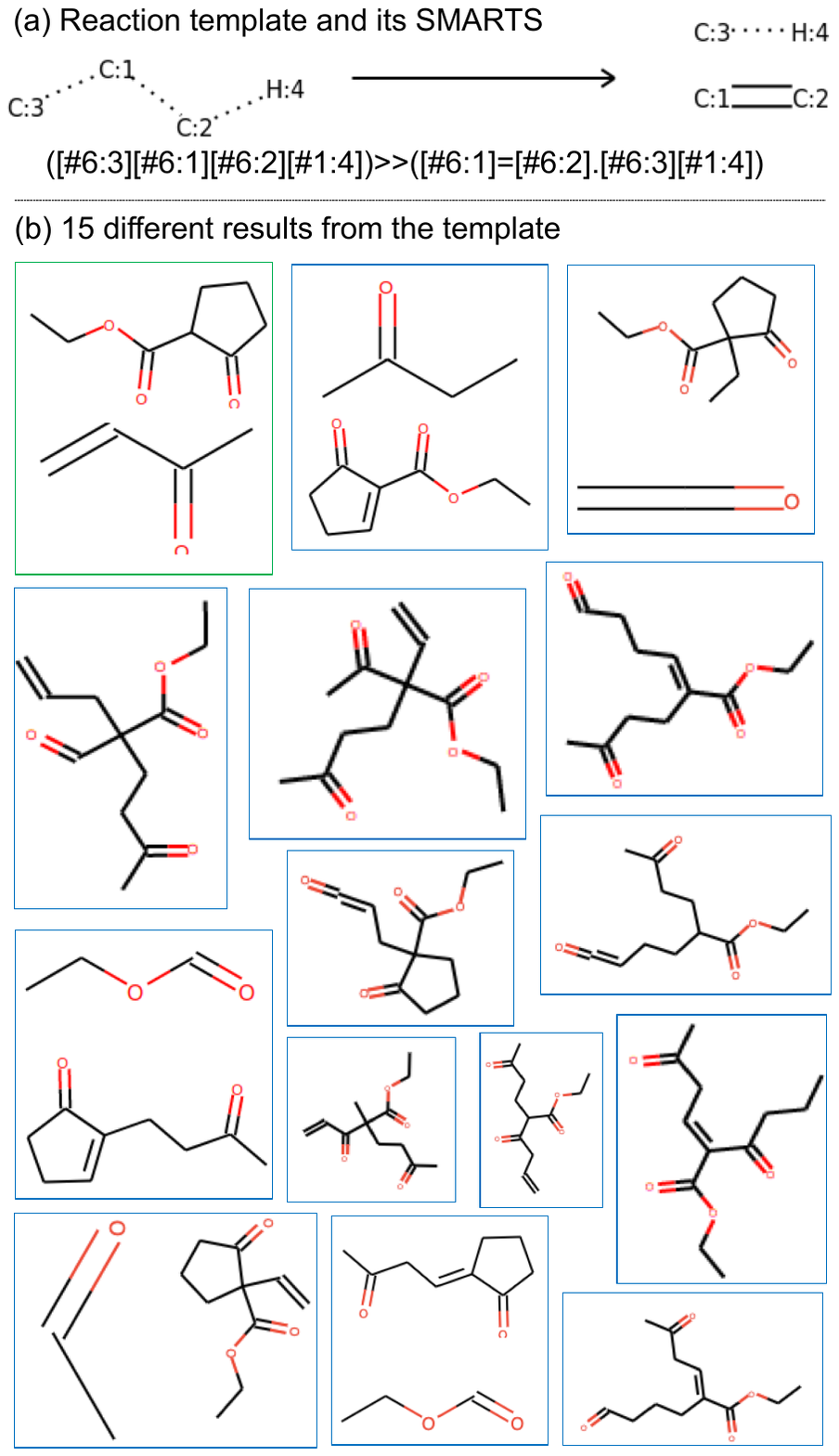}
    \caption{An example for demonstrating the application of a backward reaction template on its original product. (a) The backward template derived from the reaction center (n=0) shown in Fig.~\ref{fig:reaction_template_ex}, together with its SMARTS string. In the SMARTS, there are 4 atom-atom mappings between two sides of '$>>$' specifying the transformation of atoms in the reaction center. (b) 15 different results from applying the template back on the original product, with the original reactants in green box. More general templates lead to more possible results, implying more promiscuous enzymes.}
    \label{fig:template_ex1}
\end{figure}
\begin{figure}
\centering
    \includegraphics[width=\textwidth, trim={2cm, 8.8cm, 2cm, 0.5cm}, clip]{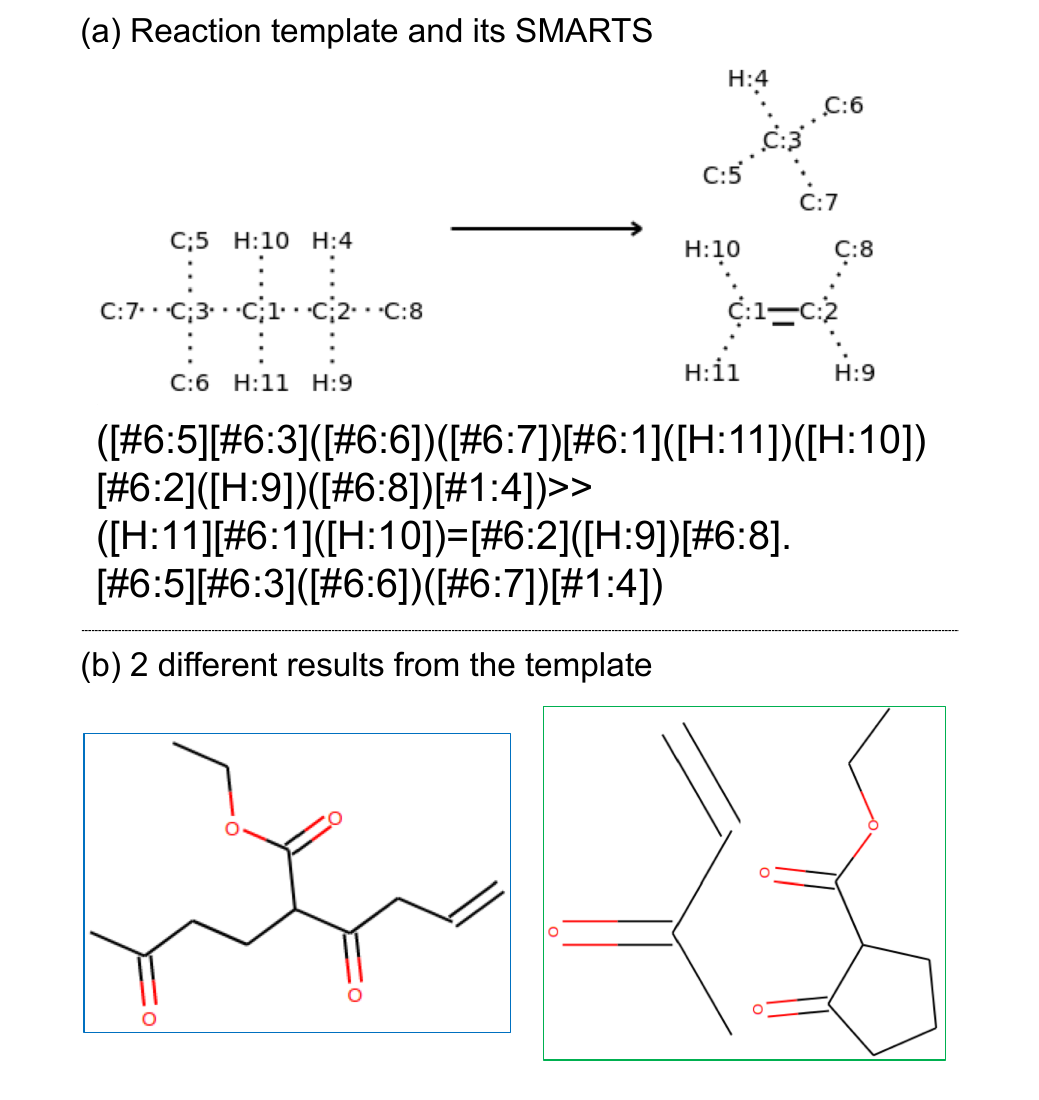}
    \caption{An example demonstrating the application of a more specific backward reaction template on its original product. (a) The backward template derived from the n=1 template shown in Fig.~\ref{fig:reaction_template_ex}, together with its SMARTS string. In the SMARTS, there are 11 atom-atom mappings between two sides of '$>>$', specifying the transformation of  more atoms not just in the reaction center as in Fig.~\ref{fig:template_ex1}. (b) two different results from applying the template back on the original product, with the original reactants inside green box. More restricted templates lead to fewer possible results, implying more specific enzymes.}
    \label{fig:template_ex2}
\end{figure}
A reaction template is an abstract description of a reaction \citep{lin2019retrosynthetic}. 
Reaction centers are those atoms and bonds that change their connections with their neighbors in a reaction.
The reaction center is a general (or promiscuous, in biological terms) reaction template since it does not consider the impact of any other atoms in the reaction. 
Other atoms that are around the reaction center can be incorporated into reaction center to form a more specific reaction template. 
Hence, depending on constraints and specifications, one reaction can have many templates.
As shown in Fig.~\ref{fig:reaction_template_ex}, by simply varying the distance to the reaction center, one can change which atoms are included in a template. Increasing distance generates more specific templates. In this sense, the original reaction is also a reaction template of itself. 
Backward reaction templates can be derived easily by switching the place of reactant(s) and product(s), namely, the two sides of '$>>$', in forward templates. 
The corresponding backward templates for $n=0$ and $n=1$ templates in Fig.~\ref{fig:reaction_template_ex}, together with their SMARTS strings, are shown in Fig.~\ref{fig:template_ex1}(a) and Fig.~\ref{fig:template_ex2}(a), respectively.
Four AAMs are specified in the $n=0$ backward template (Fig.~\ref{fig:template_ex1}(a)), while eleven AAMs are specified for the $n=1$ template (Fig.~\ref{fig:template_ex2}(a)).

\textbf{Computational tools.}
All scripts were written in Python. 
RDKit was used for molecule/reaction manipulation and various cheminformatics calculations \citep{rdkit}. 
Keras \citep{geron2019hands} using the Tensorflow 2 \citep{abadi2016tensorflow} backend was used for building the machine learning architecture.

\subsection{Templates selection}
\begin{figure}
\centering
    \includegraphics[width=0.95\textwidth, trim={2.7cm, 6.1cm, 2.5cm, 1.1cm}, clip]{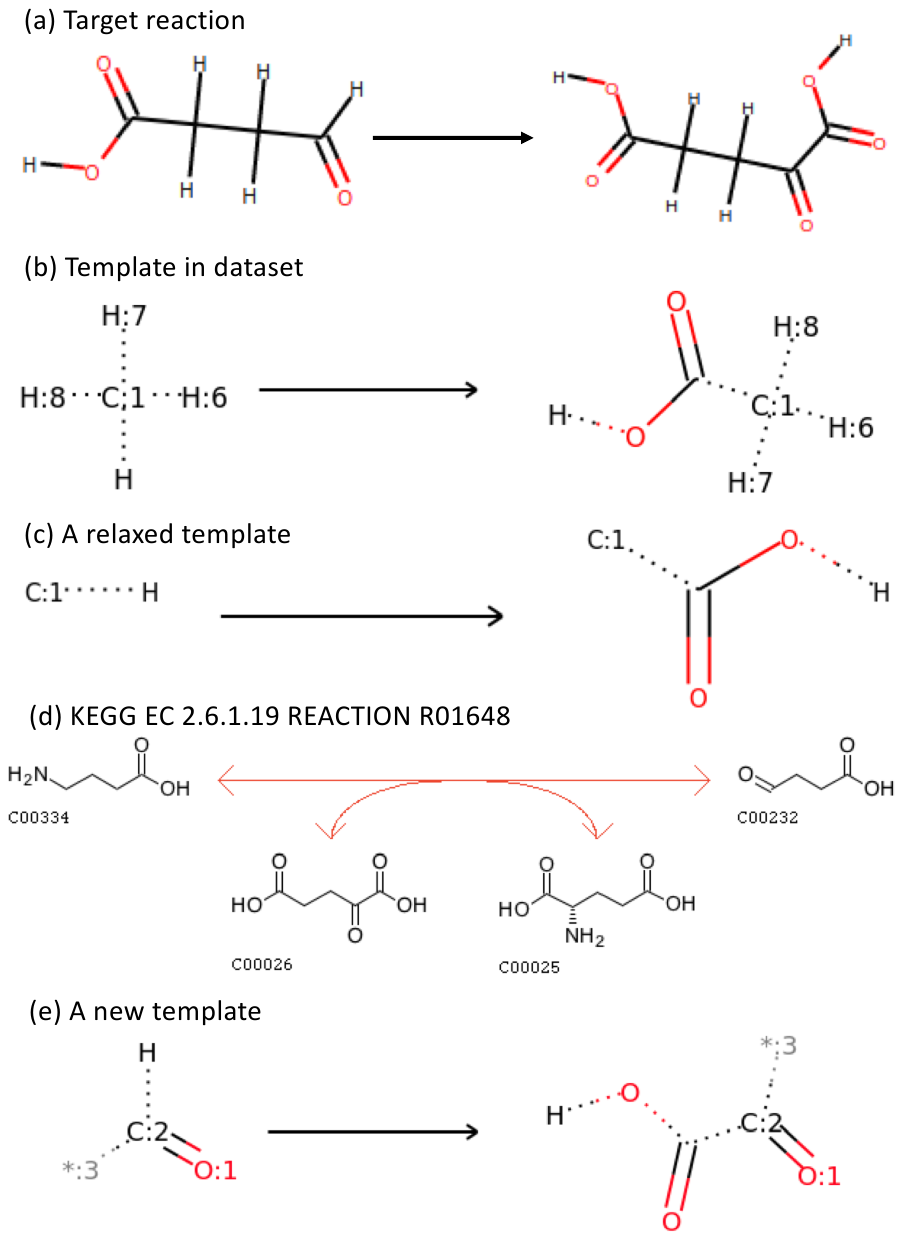}
    \caption{Demonstrating template selection with a reaction in the BDO pathway. (a) The target reaction which was step 1 in the BDO pathway (Fig.~\ref{fig:bdo_pathway}). (b) A template in the original dataset that carried out similar enzymatic transformation as the case in (a). (c) A template derived by relaxing constraints on carbon 1 in (b). This template made the transformation in (a) possible but the template was very promiscuous. (d) An existing reaction $R01648$ with EC $2.6.1.19$ in KEGG that has similar enzymatic transformation as in (a). (e) A new template derived from (d) with the carbon atom more well-constrained than in (c).}
    \label{fig:template_selection}
\end{figure}
Template-based methods are all dependent on templates that are included in a system \citep{delepine2018retropath2, lin2019retrosynthetic}. 
As discussed in many papers \citep{lin2019retrosynthetic, henry2010discovery}, there is no perfect template set. 
The selection of templates is up to the users to best suit their applications. 
When encountering a capability issue in a pipeline, namely, some enzymatic transformation cannot be carried out, one should consider adding more templates as illustrated by the case when we were trying to reproduce the BDO pathway reported by Yim et al.\ (Fig.~\ref{fig:template_selection}) \citep{yim2011metabolic}. 
The pipeline initially failed to reproduce step 1 in the BDO pathway because the template in the dataset carrying a similar enzymatic transformation had a strong constraint on its carbon atom, thus preventing it from matching with the target substrate (Fig.~\ref{fig:template_selection} (a) and (b)). 
There were two ways of rescuing this functionality issue. 
One was by relaxing the constraint on the carbon atom to get a more general template, but the resulting template was promiscuous since there were almost no constraints on the carbon atom (Fig.~\ref{fig:template_selection} (c)).
A second way was to check some metabolic reaction databases like KEGG for existing reactions having similar enzymatic transformation, and then derive new template(s) from those reactions. 
We did find reaction R01648 with EC 2.6.1.19 in KEGG that served our purpose and derived a new template that fixed the issue (Fig.~\ref{fig:template_selection} (d) and (e)). 
The capability of the framework depended heavily on the coverage of the template set.
A more diverse set of backward reaction templates would allow the framework to produce more varied candidates, improving its capability for different kinds of pathway design.

\subsection{Generating precursors in pathway design}
\textbf{Applying a template on a target.} 
Applying a reaction template on a target --- either a single molecule or multiple molecules --- computationally means finding substructures in the target that match exactly with what is specified in the template, then applying the transformation to change the structure of the target \citep{daylightsmarts}.
Hence, applying a forward (or backward) template derived from a reaction on its original reactants (or products) may produce more than one result if multiple matched substructures exist. 
As shown in Fig.~\ref{fig:template_ex1}, for a template containing only the reaction center, there are not many constraints on the targeted substructure (a three-carbon backbone with a hydrogen on one side in this example), hence there are many such substructures in the target molecule (the product in the 1st row of Fig.~\ref{fig:reaction_template_ex}), leading to 15 different results. Some computationally generated results are not valid molecular structures, but they demonstrate the potential behaviors of such loose templates as mimicking promiscuous enzymes in nature.
The number of outcomes from applying a template could be reduced, sometimes significantly, if more information is incorporated into the template. 
As shown in Fig.~\ref{fig:template_ex2}(a), the same three-carbon backbone is now fully constrained by neighbouring atoms.
With these constraints, especially constraints on atom 3, only one carbon atom in the target molecule could match with atom 3, and then only two carbon atoms match with atom 1 which is constrained by two hydrogen atoms. Hence, only 2 substructures in the target molecule match what is specified in the template, leading to only 2 possible outcomes in Fig.~\ref{fig:template_ex2}(b).
This kind of behavior imitates enzyme specificity, namely, more complicated templates could reduce the number of matched substructures, thus reducing the number of targets - substrates in biochemistry - that can be modified by a template. 
In short, deriving a template from a given reaction in some way also settles the trade-off between enzyme specificity and promiscuity.

One would expect that using a more complicated template would reduce running time, since it usually produces fewer outcomes as shown above.
However, in reality this is not always the case. Recall that applying a template is actually a process of searching for substructures in a graph representing the target. 
If some atoms in the specified template are interchangeable, switching those atoms will yield more matched substructures computationally, and hence taking extra computational time, even though producing repeating outcomes as is the case in Fig.~\ref{fig:template_ex2}(a).
In this case, hydrogen atoms 10 and 11 are interchangeable, hydrogen atoms 4 and 9 are interchangeable, and carbon atoms 5, 6, 7 are interchangeable. Three interchangeable carbon atoms bring in 6 (by $3\times2$) permutations, and two interchangeable carbon atoms bring in 2 permutations. 
Hence, each matched substructure actually has $6\times2\times2=24$ permutations, and totally 48 outcomes were produced in this case. 
In running time measured with the \texttt{time} package in Python, it took 0.127 seconds per 100 repeats for the simple case (Fig.~\ref{fig:template_ex1}), and 0.188 seconds per 100 repeats for the complex one (Fig.~\ref{fig:template_ex2}), although both have the same amortized running time of 0.004 seconds per 100 repeats per outcome. 
Note that the amortized running time is similar here primarily because two substructures are not very different from each other. It could vary if matched substructures are very different, since substructure search on a graph is an NP problem. 

For software implementation in \texttt{RDKit}, given a target and a template, the procedure is: 
\begin{enumerate}
    \item Load and sanitize the given target, make hydrogen explicit with function \texttt{Chem.AddHs()} if needed. 
    \item Run the reaction with function \texttt{RunReactants()}. 
    \item Sanitize results with function \texttt{Chem.SanitizeMol()}.
    \item Remove duplicates. 
\end{enumerate}

\subsection{Pathway ranking} \label{sec:SI-PR}
Once precursor candidates were generated, the following step was ranking those candidates so as to only expand top-ranked candidates for further explorations. 
Many quantitative analyses could be used for this task, but molecular structure based methods are popular due to their simplicity since no further experimental data were required \citep{coley2017prediction}. 
There are two key elements in developing a structure-based ranking method, one is a mathematical representation of the relevant molecular information and the other is a plausibility metric measuring similarity between given starting compound and generated precursor(s) and/or probability of a hypothetical reaction being a valid reaction \citep{maggiora2004molecular, maggiora2011molecular, maggiora2014molecular}. 

\textbf{Molecular fingerprint as a mathematical representation of molecule(s).} 
Molecular information could be represented in many mathematical forms such as graphs, sets, vectors and functions \citep{maggiora2011molecular}. 
Molecular fingerprint is a typical way of representing molecules where each component corresponds to a "local" or "global" feature or property of a molecule, for instance, molecular fragments/substructures, molecular weight, logP, etc.. \citep{maggiora2011molecular}.
For binary-valued feature vectors, their components usually indicate the presence or absence of a feature. Hence, they are also referred to as bit vectors. 
For integer- and categorical-valued feature vectors, their components could refer to the frequency of occurrence of a given feature. 
Feature vectors could also consist of continuous values, be learnt through neural networks, and many more \citep{maggiora2014molecular, rogers2010extended, duvenaud2015convolutional, riniker2013open, duan2010analysis, kearnes2016molecular, coley2017convolutional}.

\textbf{Similarity metrics.}
The most widely used similarity measure by far is the Tanimoto similarity coefficient $S_{Tan}$ which is symmetric and given by \citep{maggiora2011molecular} 
\begin{equation}
    S_{Tan}(A,B) = \frac{|A\cap B|}{|A\cap B| + |A-B| + |B-A|}
\end{equation}
in set-theoretic language, and as \citep{coley2017computer}
\begin{equation}
    S_{Tan}(A,B) = \frac{\sum A_i B_i}{\sum A_i^2 + \sum B_i^2 - \sum A_i B_i}
\end{equation}
if $A$ and $B$ are given in vector form \citep{tanimoto1958elementary}. It represents the ratio of the number of common features over the number of all unique features in two subjects being compared.
Tversky generalized the typical symmetric Tanimoto measure and defined an asymmetric similarity measure $S_{Tve}$ as 
\begin{equation}
    S_{Tve}(A,B) = \frac{|A\cap B|}{|A\cap B| + \alpha|A-B| + \beta|B-A|}
\end{equation}
where $\alpha$ and $\beta$ are both non-negative real value \citep{tversky1977features}. It reduces to Tanimoto if $\alpha$ and $\beta$ both are $1$. While for all other value combinations of $\alpha$ and $\beta$, $S_{Tve}(A,B)$ is asymmetric. 
One well-known limitation of using these similarity metrics is that they usually work again the selection of larger molecules in the search which wouldn't be desired in metabolic pathway design if larger target molecules are biologically active but could be missed by similarity metrics \citep{maggiora2011molecular, maggiora2014molecular}. 

\textbf{Machine learning for pathway ranking.}
Traditional machine learning methods, like support vector machine \citep{delepine2018retropath2}, have been incorporated into ranking models. But the application of deep learning as ranking methods only took off a few years ago, following the boom of using deep learning for knowledge representation \citep{lecun2015deep, coley2017prediction}.
In this work, we explored the possibility of using deep learning to guide metabolic pathway design.

\subsection{Additional details of proposed approach} \label{sec:SI-methods}

In the work, we developed a new framework for \textit{de novo} metabolic pathway design by leveraging cheminformatics tools and deep learning. 
Given a target compound of interest for production through metabolic engineering, one has to first design a metabolic pathway that connects the target to some available metabolite(s) in chassis before any further experiments could be carried out. 
The basic idea here was to use backward reaction template-based enumeration to generate possible precursor candidates and use neural network based ranking models to pick out high-rank candidates for next step expansion.   
Deep-learning based ranking models were trained on KEGG metabolic reactions to learn the underlying connections between metabolic reactants and products. 
Then the probability score generated by the trained neural network-based pathway ranking models (NN1PR/NN2PR) given any reactant-product pair was used for pruning and ranking all candidates. 
As depicted in Fig.~\ref{fig:multistep_pipeline}, given a target compound, for the first-step reaction, some backward reaction templates were applied to generate potential candidates that were represented as three light orange circles in the figure. 
Then NN1PR assigned each candidate a score in unit interval, which was used to rank those candidates. 
Depending on computational time requirement or user's settings, only a portion of high-ranked candidates would be chosen for designing the second-step reaction. 
In the illustrating schematic, only the first 2 candidates were chosen for next step expansion which yielded 6 possible candidates as represented by green circles. 
And again, NN1PR was used to prune some candidates (2 in the example) and rank the rest (ranks 1 to 4 in the figure). The process went on and on until finishing preset number of steps.  
Generated pathways would then be analyzed to see if any could lead to known cellular metabolites. If that is the case, then at least one synthetic pathway can be reconstructed through backtracking.

\subsubsection{Data preprocessing}

\textbf{Querying KEGG for SMILES of each compound in metabolic reaction dataset.} 
As mentioned in Section Dataset assembly, compounds in the metabolic reaction dataset were in KEGG compound IDs, which follow KEGG nomenclature but do not contain any cheminformatic information directly. 
To get SMILES of each compound, we queried KEGG for its compound list which had 18749 entries in total.
We set up a pipeline in Python to automatically query KEGG for the SMILES string of each KEGG compound listed. 
It took about 378 min to complete the querying pipeline, approximately 2.1 seconds per query. 
Among those 18749 KEGG IDs, 665 of them did not have corresponding SMILES strings in the KEGG database (marked as KEGG error compounds), and another 26 returned SMILES could not be processed by RDKit (marked as RDKit error compounds).
Upon analyzing all SMILES strings we got, we found that there were 17751 unique SMILES strings out of totally 18058 KEGG IDs, which clearly showed that some KEGG IDs shared the same SMILES strings. 
Further analysis showed that there were 188 SMILES strings corresponding to more than one KEGG ID, and some SMILES strings were shared by as many as 16 entries. 
Specifically, there were 141 SMILES strings having 2 KEGG IDs, 27 having 3 IDs, 7 having 4 IDs, 6 having 5 ID, and another 8 SMILES strings having 6, 8, 9, 10, 11, 14, and 16 IDs respectively. 
In summary, most compounds sharing SMILES strings were generic compounds in KEGG. 
Some biologically different compounds couldn't be distinguished by SMILES as SMILES couldn't capture 2D/3D information, leading those compounds to have the same SMILES. 


\textbf{Data cleaning.}
For 26 RDKit error compounds, we manually checked each compound's KEGG page and found that they were all generic compounds in reaction hierarchy which used some KEGG notations to denote generic atoms in MolFile. Thus, RDKit failed to process those MolFiles because those generic atom symbols were not real atom symbols. 
So we did two replacements manually as a temporary fix to this issue. 
One was replacing atom symbol 'X' with 'Cl' even though in KEGG and chemistry, it stands for any halogen elements ('F', 'Cl', 'Br', 'I'). 
Another was to replace alkyl group symbol 'R' with methyl group 'C'. 
These two replacements would fix generic reactions into specific reactions, and hence maybe limiting the capability of any work using this dataset. 
10 RDKit error compound were fixed by the method above, while the others
('cpd:C01041', 'cpd:C02202', 'cpd:C12862', 'cpd:C13681', 'cpd:C13932', 'cpd:C18368', 'cpd:C18380', 'cpd:C18384', 'cpd:C19040', 'cpd:C19600', 'cpd:C20442', 'cpd:C21011', 'cpd:C21012', 'cpd:C21013', 'cpd:C21014', 'cpd:C22197') couldn't be fixed and would be removed from our dataset. 
A better approach to fixing the generic compounds issue would be to replace each generic compound with a set of specific compounds, thereby expanding each generic reaction into multiple reactions. In this way, one could increase the size of the dataset without much effort. 
But the expanded dataset would not improve the coverage of the dataset in terms of reaction types, namely, no benefit to dataset diversity. 
Given our purpose as proof-of-concept study, we expected the impact to be very limited. 
Since we were using this dataset for training and validation of a machine learning model, the diversity and coverage of the dataset were more important than its size. 
Our handling did preserve dataset diversity, and hence would not impact results very much. 

For 665 KEGG error compounds, there seemed no easy way to fix them. So we chose to remove them from the metabolic reaction dataset.
For each reaction, its reactants and products were checked so as to be clean of KEGG error compounds. 
If either a reaction's reactant part or product part or both contained no compounds after cleaning, that reaction was tagged as broken and was removed from the dataset. 
After this cleaning step, only one KEGG reaction `R06748' was removed and we obtained a dataset with 6649 metabolic reactions and 6026 compounds. 
We understood that this cleaning operation would make balanced reactions unbalanced. 
We observed that, unlike chemical reaction databases, most metabolic reactions in metabolic reaction databases were not atom-balanced. 
The focus of metabolic reactions usually was on key substrates and products. So our cleaning step was in line with this observation and our pipeline here was also going to focus on key participants of metabolic reactions.   

\textbf{Process reactions into mono-product reactions.}
The backward reaction templates dataset we assembled were mono-product templates. 
In order to use this template dataset, we had to process our metabolic reactions dataset into mono-product reactions. 
For each multiple-product reaction, we split it into multiple children reactions sharing the same reactants but each just inherited one product from their parent reaction.   
With this processing, the original 11475 reactions were expanded into 21848 single-product reactions with 7105 unique products. 
On average, each unique product occurred in 3 reactions. 
It would be better to remove a mono-product reaction if its product does not share common substructure with its reactants. This is a direction for future work. 
Examples of those compounds are usually small molecules which are usually by-products in reactions. 

\subsubsection{Data augmentation} 
\label{sec:SI-method-DA}
\begin{figure}
\centering
    \includegraphics[width=\textwidth, trim={0cm, 0cm, 0cm, 0cm}, clip]{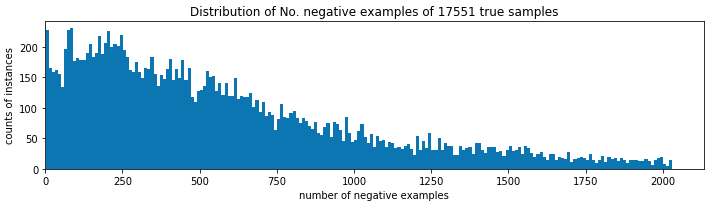}
    \caption{Distribution of number of negative examples of 17551 compounds. The number varied in a wide range from 0 to over 2000. Totally, there were $113336477$ negative reactions, about 760 for each positive one on average.}
    \label{fig:neg_ex_dist}
\end{figure}

In our metabolic reaction dataset, there were only reactions that already exist in nature (also called positive reactions in this paper).
While to build a machine learning model, one also needs negative metabolic reactions which are not favored or do not exist in nature so as to train a model to distinguish positive ones from negative counterparts. 
To create negative metabolic reactions, we adapted a data augmentation procedure reported by Coley et al.\ \citep{coley2017prediction}.
We applied all reaction templates - both forward and backward - on all unique SMILES strings, totally 18058 cases, to generate hypothetical precursors for each compound. 
This augmentation process --- applying hundreds of thousands of reaction templates on thousands of targets one-by-one --- was time consuming, taking several hours in our implementation. 
For each mono-product reaction, the set of hypothetical precursors of its product was used to generated negative reactions by simply linking each hypothetical precursor with the product by `$>>$' to form a valid reaction SMILES. 
The original precursors were removed from hypothetical precursors beforehand. 
To guarantee that all hypothetical precursors were hypothetical but not found in nature, a strict check was carried out to remove all positive cases from generated counterparts. 
There were 200 compounds that couldn't be modified by any templates, and most of them were small molecules. 
After this augmentation step, some compounds had more than 2000 augmented precursor sets, but it was about 760 cases for each compound on average (Fig.~\ref{fig:neg_ex_dist}).
The statistics here signaled the importance of a well-performed ranking model which would be tasked to rank the positive ones as high as possible among hundreds of candidates.
Overall, we got $113336477$ negative reactions.

\subsubsection{Reaction fingerprints}
\label{sec:SI-methods-RF}
Most of existing machine learning models require input data to be in tensor format \citep{geron2019hands, goodfellow2016deep, francois2017deep}, so our datasets in SMILES and SMARTS had to be vectorized for using machine learning model. 
As briefly mentioned in Section Manipulation molecules in cheminformatics, molecular fingerprints \citep{todeschini2008handbook}, such as Extended-connectivity fingerprints (ECFPs) \citep{rogers2010extended} that were derived from the Morgan Algorithm \citep{morgan1965generation} and learned fingerprints \citep{duvenaud2015convolutional, kearnes2016molecular},   have been widely used to convert molecules in SMILES into vectors.
Some researchers have developed machine learning models that use SMILES/SMARTS sequences as input directly \citep{liu2017retrosynthetic}.
Reaction fingerprints can be derived from the concatenation of reactant and product fingerprints \citep{wei2016neural}, the difference between reactant and product fingerprints \citep{kraut2013algorithm}, or many other ways \citep{kayala2011learning, kayala2012reactionpredictor}. 
In this work, we used an implementation of ECPFs in RDKit to generate molecular fingerprints, and represented each reaction as the concatenation of its reactant and product fingerprints. Each group of reactants or products was represented as a 512-bit vector and hence each reaction was a 1024-bit vector. Each bit represented a feature and took value 1 if corresponding feature existed, otherwise 0.

\subsubsection{Deep learning model}
\paragraph{Neural network based one-step pathway ranking model (NN1PR).}
The dataset had $113336477$ negative examples but only $21848$ positive counterparts, which was highly unbalanced. For a balanced training, we first random shuffled negative examples, and only used top $1564651$ negative examples (30\%) for training. 
The deep learning model was a feedforward neural network consisting of one hidden dense layer with `relu' as the activation function \citep{goodfellow2016deep} (Fig.~\ref{fig:model_architectures}a). 
The hidden layer had 256 neurons, followed by a dropout layer to regulate overfitting.
The output layer was a dense layer with 1 neuron using 'sigmoid' as the activation function, so as to represent the probability of a reaction being a positive one. 
The input layer was a trivial one that didn't have any parameters.

\begin{figure}
\centering
    \includegraphics[width=\textwidth, trim={0cm, 0cm, 0cm, 0cm}, clip]{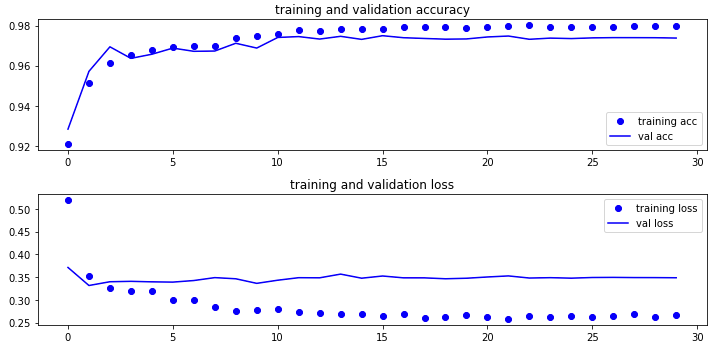}
    \caption{Training progress of the deep learning model. The model was trained for 30 epochs with a batch size of 128. The performance based on either loss or classification accuracy improved in the first 10 epochs and then plateaued afterwards.}
    \label{fig:ml_training}
\end{figure}

Mathematically, the hidden layer was
$$\bm{h}^{(1)} = relu^{(1)}(\bm{W}^{(1)T}\bm{x} + \bm{b}^{(1)})$$
where $\bm{W}^{(1)} \in \mathbb{R}^{1024\times 256}$ and $\bm{b}^{(1)} \in \mathbb{R}^{256\times 1}$.
The output layer used a sigmoid function as the activation function, and was given by 
\begin{align*}
    &{h}^{(3)} = \bm{W}^{(3)T}\bm{h}^{(1)} + {b}^{(3)},  \\
    &sigmoid(h^{(3)}) = \frac{1}{1 + exp(-h^{(3)})}
\end{align*}
where $\bm{W}^{(3)} \in \mathbb{R}^{256\times 1}$ and ${b}^{(3)} \in \mathbb{R}$.
The mathematical representation of the dropout layer was not presented here. 
The hidden layers had 262400 parameters, and the output layer had 257 parameters. 
So taken together, this model had $262657$ parameters that were all going to be trained in the training process. 
The optimizer was Adam, which is a stochastic gradient descent method that is based on adaptive estimation of first-order and second-order moments, with a learning rate as 0.001 \citep{kingma2014adam}.
The loss function was the built-in 'binary\_crossentropy' in TensorFlow and during the training, we also monitored classification accuracy. 
The model was trained for 50 epochs with a batch size of 128 (Fig.~\ref{fig:ml_training}). 
The performance based on either loss or classification accuracy improved in the first 10 epochs and then plateaued afterwards. 
Based on this observation, we determined that 30 epochs would be a good place to stop the training.
We observed that the training curves varied slightly at the beginning in different runs, but all reached similar final performance. The difference could be explained by the random initialization of model parameters.

\begin{figure}
\centering
    \includegraphics[width=0.7\textwidth, trim={0cm, 0cm, 0cm, 0cm}, clip]{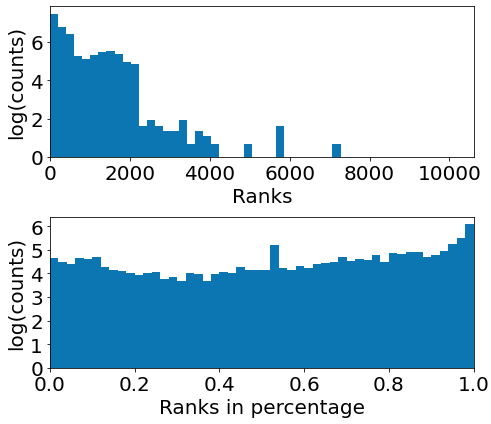}
    \caption{Performance of baseline model on ranking testing samples. 
    Top: the distribution of ranks of 4796 testing samples by the baseline.
    Bottom: the distribution of ranks in percentage of all testing samples by the baseline.}
    \label{fig:ml_dd_bl}
\end{figure}
\begin{figure}
\centering
    \includegraphics[width=0.7\textwidth, trim={0cm, 0cm, 0cm, 0.cm}, clip]{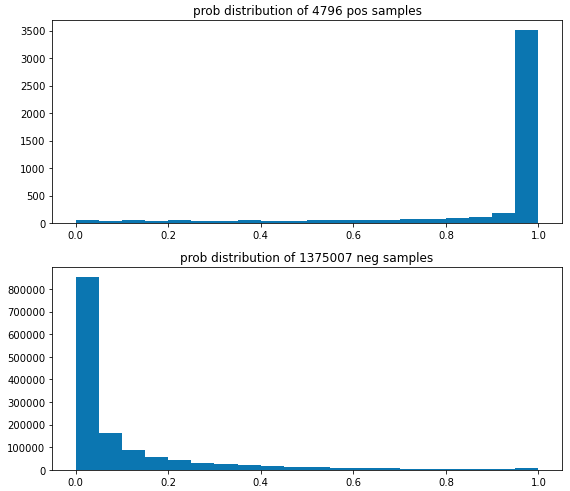}
    \caption{Performance of NN1PR on reaction classification. 
    Top: probability distribution of 4796 positive testing cases. 
    Bottom: probability distribution of 1375007 negative testing cases. 
    Probabilities were evaluated by NN1PR.}
    \label{fig:ml_dd_ml}
\end{figure}
\begin{figure}
\centering
    \includegraphics[width=0.8\textwidth, trim={0cm, 0cm, 0cm, 0cm}, clip]{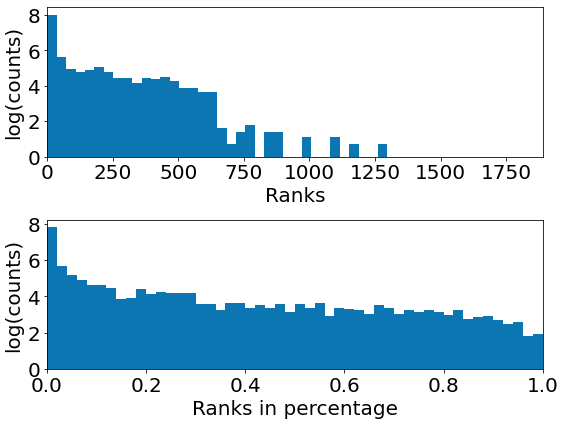}
    \caption{Performance of NN1PR on ranking testing samples. 
    Top: the distribution of ranks of 4796 testing samples by NN1PR.
    Bottom: the distribution of ranks in percentage of all testing samples by NN1PR.}
    \label{fig:ml_performance}
\end{figure}

NN1PR was trained as a binary classifier, so we tested its capability of classifying reactions on reserved positive reactions and other unused negative reactions (Fig.~\ref{fig:ml_dd_ml}). 
The probability distributions showed that NN1PR was able to assign low scores to most negative reactions while most positive ones received high scores.  
Both the baseline model and the machine learning model were then used to rank positive reactions together with negative reactions.
Each positive reaction was mixed with its negative counterparts and two models were used to rank all reactions. 
For the machine learning model, most cases were ranked within 700 (Fig.~\ref{fig:ml_performance}), and almost all positive reactions had ranks within 2000.
For the baseline model, some reactions were ranked higher than $10000$ (Fig.~\ref{fig:ml_dd_bl}).
The machine learning model outperformed the baseline model (Fig.~\ref{fig:model_performance}a). More than 50\% of samples were ranked in the top-10 by the deep learning model and more than 70\% were within rank 100, while for the baseline, only about 23\% were within the top-100.

\paragraph{Neural network based two-step pathway ranking model (NN2PR).}
The pathway dataset had $1280074$ negative examples but only $1420$ positive counterparts, which was highly unbalanced. 
For a balanced training, we assigned a weight as the ratio between the number of negative and positive samples to each positive reaction. 
The deep learning model was a feedforward neural network consisting of two hidden dense layers with `relu' as the activation function \citep{goodfellow2016deep} (Fig.~\ref{fig:model_architectures}b). 
The hidden layers, respectively, had 512 and 128 neurons, each followed by a dropout layer to regulate overfitting.
The output layer was a dense layer with 1 neuron using 'sigmoid' as the activation function, so as to represent the probability of a 2-step pathway being a positive one. 
The input layer was a trivial one that didn't have any parameters.

\begin{figure}
\centering
    \includegraphics[width=\textwidth, trim={0cm, 0cm, 0cm, 0cm}, clip]{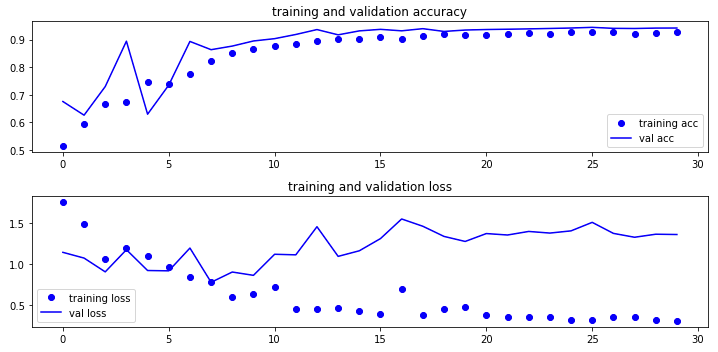}
    \caption{Training progress of NN2PR. The model was trained for 30 epochs with a batch size of 128. The performance based on either loss or classification accuracy improved in the first 15 epochs and then plateaued afterwards.}
    \label{fig:nn2pr_training}
\end{figure}

Mathematically, the first hidden layer was
$$\bm{h}^{(1)} = relu^{(1)}(\bm{W}^{(1)T}\bm{x} + \bm{b}^{(1)})$$
where $\bm{W}^{(1)} \in \mathbb{R}^{1536\times 512}$ and $\bm{b}^{(1)} \in \mathbb{R}^{512\times 1}$.
The second hidden layer was given by 
$$\bm{h}^{(2)} = relu^{(2)}(\bm{W}^{(2)T}\bm{h}^{(1)} + \bm{b}^{(2)})$$
where $\bm{W}^{(2)} \in \mathbb{R}^{512\times 128}$ and $\bm{b}^{(2)} \in \mathbb{R}^{128\times 1}$.
The output layer used a sigmoid function as the activation function, and was given by 
\begin{align*}
    &{h}^{(3)} = \bm{W}^{(3)T}\bm{h}^{(2)} + {b}^{(3)},  \\
    &sigmoid(h^{(3)}) = \frac{1}{1 + exp(-h^{(3)})}
\end{align*}
where $\bm{W}^{(3)} \in \mathbb{R}^{128\times 1}$ and ${b}^{(3)} \in \mathbb{R}$.
The mathematical representation of the dropout layer was not presented here. 
The hidden layers had 786944 and 65664 parameters, separately, and the output layer had 129 parameters. 
So taken together, this model had $852737$ parameters that were all going to be trained in the training process. 
The optimizer was Adam, which is a stochastic gradient descent method that is based on adaptive estimation of first-order and second-order moments, with a learning rate as 0.001 \citep{kingma2014adam}.
The loss function was the built-in 'binary\_crossentropy' in TensorFlow and during the training, we also monitored classification accuracy. 
The model was trained for 50 epochs with a batch size of 128 (Fig.~\ref{fig:nn2pr_training}). 
The performance based on either loss or classification accuracy improved in the first 15 epochs and then plateaued afterwards. 
Based on this observation, we determined that 30 epochs would be a good place to stop the training.
As with NN1PR, we observed that the training curves varied slightly at the beginning in different runs, but all reached similar final performance.

\begin{figure}
\centering
    \includegraphics[width=0.7\textwidth, trim={0cm, 0cm, 0cm, 0cm}, clip]{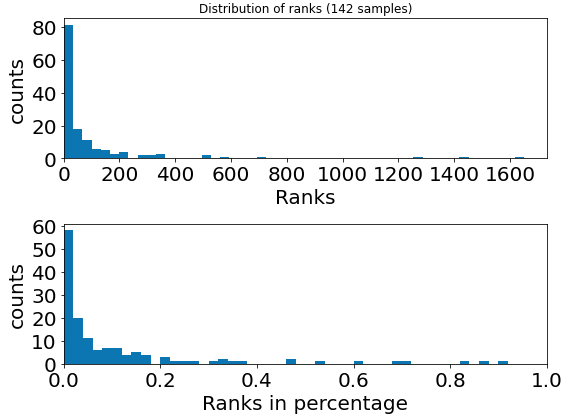}
    \caption{Performance of NN1PR on ranking 2-step testing pathways. 
    Top: the distribution of ranks of 142 testing samples by NN1PR.
    Bottom: the distribution of ranks in percentage of all testing samples by NN1PR.}
    \label{fig:nn2pr_baseline}
\end{figure}
\begin{figure}
\centering
    \includegraphics[width=0.7\textwidth, trim={0cm, 0cm, 0cm, 0.cm}, clip]{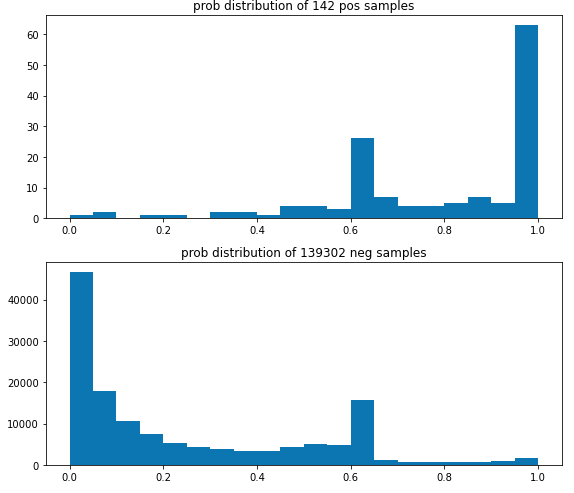}
    \caption{Performance of NN2PR on reaction classification. 
    Top: probability distribution of 142 positive testing cases. 
    Bottom: probability distribution of 139302 negative testing cases. 
    Probabilities were evaluated by NN2PR.}
    \label{fig:nn2pr_classification}
\end{figure}
\begin{figure}
\centering
    \includegraphics[width=0.5\textwidth, trim={0cm, 0cm, 0cm, 0cm}, clip]{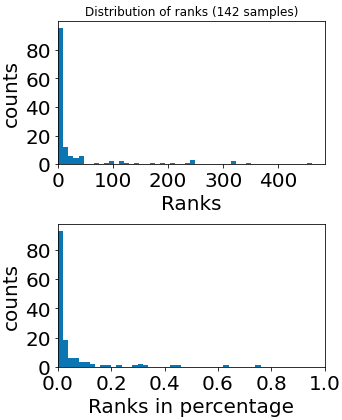}
    \caption{Performance of NN2PR on ranking testing samples. 
    Top: the distribution of ranks of 142 testing samples by NN2PR.
    Bottom: the distribution of ranks in percentage of all testing samples by NN2PR.}
    \label{fig:nn2pr_ranks}
\end{figure}

NN2PR was trained as a binary classifier, so we tested its capability of classifying reactions on reserved positive reactions and other unused negative reactions (Fig.~\ref{fig:nn2pr_classification}). 
The probability distributions showed that NN2PR was able to assign low scores to most negative reactions while most positive ones received high scores, although there were many cases that could not be clearly classified.
Both the baseline model and the machine learning model were then used to rank positive 2-step pathways together with negative counterparts.
Each positive pathway was mixed with its negative counterparts and two models were used to rank all cases.
NN2PR ranked most cases in the top-50 (Fig.~\ref{fig:nn2pr_ranks}), and almost all positive reactions had ranks within 400, while for the baseline model, some reactions were ranked higher than $1600$ (Fig.~\ref{fig:nn2pr_baseline}).
NN2PR covered 25\% more cases in the top-10 than NN1PR, even though the margin was about 10\% for the top-100 (Fig.~\ref{fig:model_performance}b).

\end{document}